%% file: iclr2023_conference.tex
\documentclass{article} 
\usepackage{iclr2023_conference,times}

\input{math_commands.tex}

\usepackage{xcolor}
\usepackage{hyperref}
\usepackage{url}
\usepackage{wrapfig}
\usepackage{graphicx}
\usepackage{xspace}
\usepackage{amsmath}
\usepackage{multirow}
\usepackage{booktabs}
\usepackage{siunitx}
\usepackage[skip=8pt]{caption}
\usepackage{float}

\newcommand{\rev}[1]{\textcolor{black}{#1}}
\newcommand{\model}{\textsc{Hmr}\xspace}

\DeclareSIUnit\angstrom{\text {Å}}

\title{Learning Harmonic Molecular Representations on Riemannian Manifold}

\author{%
Yiqun Wang$^{1}$, 
\ Yuning Shen$^{1}$, 
\ Shi Chen$^{2}$\thanks{This work was conducted during internship at ByteDance Research.}, 
\ Lihao Wang$^{1}$,
\ Fei Ye$^{1}$, 
\ Hao Zhou$^{3}$\\
$^1$ByteDance Research, $^2$University of Wisconsin-Madison, \\
$^3$Institute for AI Industry Research (AIR), Tsinghua University\\
\texttt{\{yiqun.wang, yuning.shen, wanglihao.1217, yefei.joyce\}@bytedance.com}, \\
\texttt{schen636@wisc.edu},
\texttt{zhouhao@air.tsinghua.edu.cn}
}
\iclrfinalcopy 
\begin{document}
\maketitle

\begin{abstract}

Molecular representation learning plays a crucial role in AI-assisted drug discovery research.
Encoding 3D molecular structures through Euclidean neural networks 
has become the prevailing method in the geometric deep learning community.
However, the equivariance constraints and message passing in Euclidean space may limit the network expressive power.
In this work, we propose a Harmonic Molecular Representation learning (\model) framework, 
which represents a molecule using the Laplace-Beltrami eigenfunctions of its molecular surface.
\model offers a multi-resolution representation of molecular
geometric and chemical features on 2D Riemannian manifold.
We also introduce a harmonic message passing method to realize 
efficient spectral message passing over the surface manifold for better molecular encoding.
Our proposed method shows comparable predictive power to current models in small molecule property prediction,
and outperforms the state-of-the-art deep learning models for 
\rev{ligand-binding protein pocket classification and} the rigid protein docking challenge,
demonstrating its versatility in molecular representation learning.
\end{abstract}

\section{Introduction}

Molecular representation learning is a fundamental step in AI-assisted drug discovery. 
Obtaining good molecular representations is crucial for the success of downstream applications
including protein function prediction~\citep{Gl:2021structure} 
and molecular matching, e.g., protein-protein docking~\citep{Ga:2021independent}.
In general, an ideal molecular representation should well integrate both \textit{geometric} (e.g., 3D conformation) and 
\textit{chemical} information (e.g., electrostatic potential).
Additionally, such representation should capture features in various \textit{resolutions} to accommodate different tasks, 
e.g., \textit{high-level} holistic features for molecular property prediction, 
and \textit{fine-grained} features for describing whether two proteins can bind together at certain interfaces.

Recently, geometric deep learning (GDL)~\citep{Br:2017geometric,Br:2021geometric,Mo:2017geometric}
has been widely used in learning molecular representations~\citep{At:2021geometric,townshend2021geometric}. 
GDL captures necessary information by performing neural message passing~(NMP) on common structures such as 2D/3D molecular graph~\citep{Kl:2020directional, St:2020deep}, 
3D voxel~\citep{Li:2021octsurf},
and point cloud~\citep{Un:2021se}. 
Specifically, GDL encodes: a) geometric features by modeling atomic positions in the \textit{Euclidean space}, 
and b) chemical features by feeding atomic information into the message passing networks.
High-level features could then be obtained by aggregating these atom-level features, 
which has shown promising results empirically.

However, we argue that current molecular representations via NMP in the Euclidean space is not necessarily the optimal solution, 
which suffers from several drawbacks.
First, current GDL approaches need to employ equivariant networks~\citep{Th:2018tensor} to guarantee 
that the molecular representations transform accordingly upon rotation and translation~\citep{Fu:2020se}, 
which could undermine the network expressive power~\citep{Co:2018spherical, Li:2021rotation}.
Therefore, developing a representation that could properly encode 3D molecular structure 
while bypassing the equivariance requirement is desirable.
Second, current molecular representations in GDL are learned in a bottom-up manner, 
which are hardly able to provide features in different resolutions for different tasks.
Specifically, NMP in Euclidean space typically achieves long-range communication between distant atoms 
by stacking deep layers or increasing the neighborhood radius.
This would hinder the effective representation of macromolecules with tens of thousands of atoms~\citep{battiston2020networks,boguna2021network}. 
To remedy this, residue-level graph representations are commonly used for large molecules~\citep{jumper2021highly,Gl:2021structure}.
Hence designing efficient multi-resolution message passing mechanisms would be ideal for encoding molecules with distinct sizes.

\rev{
On the other hand, the molecular surface is a high-level representation of a molecule's shape,
which has been widely used to study inter-molecular interactions~\citep{richards1977areas,shulman2004recognition}.
Intuitively, the interaction between molecules is commonly described as a ``key-lock pair'', 
where both shape complementarity~\citep{Li:2013role} and chemical interactions (e.g., hydrogen bond) 
determine whether the key matches the lock molecule.
It has been shown that the molecular surface holds key information about inter-molecular interactions~\citep{Ga:2020deciphering}, 
which makes it an ideal candidate for molecular representation learning~\citep{sverrisson2021fast,somnath2021multi}.
}

\vspace{-12pt}
\begin{wrapfigure}{R}{0.55\textwidth}
\includegraphics[width = 0.55\textwidth]{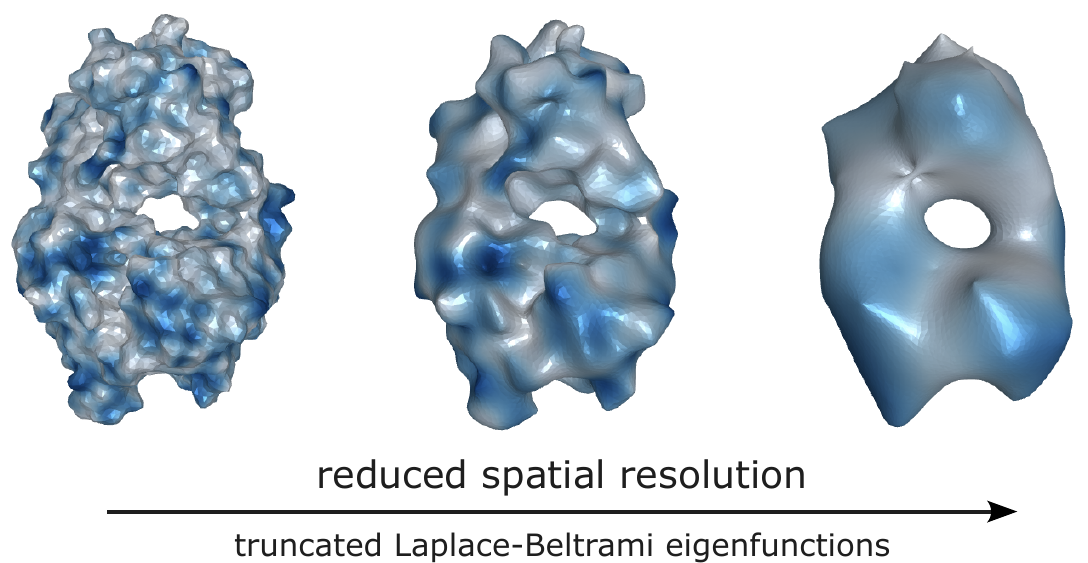}
  \caption{
  Multi-resolution molecular surface representation. Showing the electrostatic potential (blue regions being negatively charged, PDB ID: 3V6F) at different resolutions under our representation. \rev{See \autoref{ap:harmonic_analysis} for technical details about tuning resolution.}
  }
  \vspace{-8pt}
\label{fig:fig1}
\end{wrapfigure}

Inspired by the idea of \textit{Shape-DNA}~\citep{reuter2006laplace}, 
hereby we propose Harmonic Molecular Representation learning~(\model) 
by utilizing the Laplace-Beltrami eigenfunctions on the molecular surface~(a 2D manifold).
Our representation has the following advantages: 
a) \model works on 2D Riemannian manifold instead of in the 3D Euclidean space, 
thus the resulting molecular representation is by design roto-translation invariant;
b) \model represents a molecule in a top-down manner, 
and is capable of offering multi-resolution features that accommodate various target molecules 
(i.e., from small molecules to large proteins),
thanks to the smooth nature of the Laplace-Beltrami eigenfunctions (\autoref{fig:fig1});
c) \rev{\model naturally integrates geometric and chemical features 
-- the molecular shape defines the Riemannian manifold (i.e., the underlying domain equipped with a metric), 
and the atomic configurations determine the associated functions distributed on the manifold (e.g., electrostatics).
}

To demonstrate that \model is generally applicable to different downstream tasks 
including molecular property prediction and molecular matching, we propose two specific techniques: 
(1) manifold harmonic message passing for realizing holistic molecular representations, 
and (2) learning regional functional correspondence for molecular surface matching.
Without loss of generality, we apply the proposed techniques to solve three drug discovery-related problems:  
QM9 small molecule property regression, 
\rev{ligand-binding protein pocket classification,}
and rigid protein docking pose prediction.
Our proposed method shows comparable performance for small molecule property prediction to NMP-based models, 
while outperforming the state-of-the-art deep learning models in \rev{protein pocket classification and} the rigid protein docking challenge.

\section{Related Work}

\rev{
\textbf{Molecular Surface Representation}\quad
The molecular surface representation is commonly adopted for tasks involving molecular interfaces~\citep{duhovny2002efficient}, 
where non-covalent interactions (e.g., hydrophobic interactions) play a decisive role~\citep{sharp1994electrostatic}.
Non-Euclidean convolutional neural networks~\citep{Mo:2017geometric}
and point cloud-based learning models~\citep{Sv:2022physics} have been applied to encode the molecular surface for downstream applications, 
e.g., protein binding site prediction~\citep{My:2021deepsurf}.
However, existing methods apply filters with fixed sizes and are highly dependent on the surface mesh quality, 
which limit the expressive power for molecular shape representation across different spatial scales~\citep{somnath2021multi,isert2022structure}.
}

\textbf{Geometry Processing}\quad
Our work was inspired by some pioneering work for shape encoding and shape matching 
in the geometry processing research community~\citep{Li:2013learning, Bi:2016recent}.
The surface of a 3D object is typically discretized into a polygon mesh with vertices and faces.
Intrinsic properties of the surface manifold are used to encode the shape~\citep{sun2009concise,bronstein2010scale}.
Functional maps~\citep{Ov:2012functional,litany2017fully} have been proposed to establish 
spectral-space functional correspondence between two manifolds.
Recently, deep learning has been applied to learn representative features 
to facilitate shape recognition and matching~\citep{Li:2017deep, Do:2020deep, At:2021dpfm}.

\textbf{Spectral Message Passing}\quad
Our proposed method decomposes surface functions/features as the linear combination of some basis functions (hence the name ``harmonic'')
and realizes message passing by applying various spectral filters.
Graph convolutional network (GCN) is closely related to our proposed method, 
which operates in the graph Laplacian eigenspace~\citep{kipf2016semi,shen2021multi}.
The major difference is that graph is discrete by construction, 
while our method works with continuous Riemannian manifold~\citep{coifman2006diffusion}.
In other words, the underlying manifold and its spectrum remain the same with different surface discretizations, 
hence is a robust representation of the surface shape~\citep{coifman2005geometric}. See ``Shape-DNA'' in \autoref{sec:preliminaries}.

\section{Preliminaries}\label{sec:preliminaries}

The goal of this work is to propose a representation learning method using the molecular surface, 
which could properly encode both geometric and chemical features.
Intuitively, the molecular surface defines the shape of a molecule in 3D Euclidean space (i.e., geometry), 
while chemical features (e.g., hydrophobicity) can be treated as functions distributed on the molecular surface.
The surface and these associated functions co-determine the underlying molecular properties, 
e.g., whether an antibody could bind with an antigen.
Therefore, we propose to represent a molecule as a set of (learned) functions/features defined on its surface.

\begin{figure}[h]
\centering
\vspace{6pt}
\includegraphics[width=0.98\textwidth]{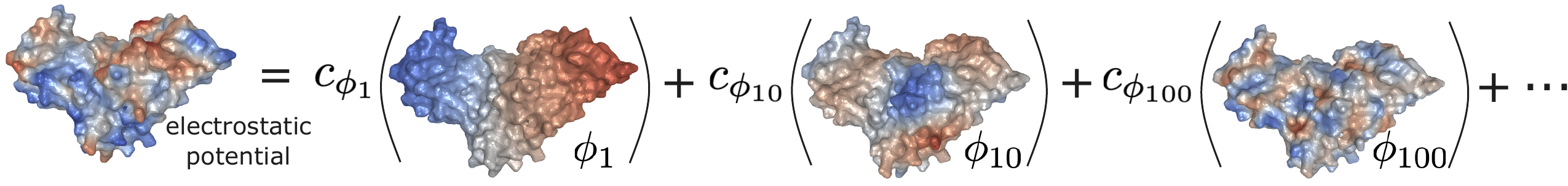}
  \caption{
  Illustrating manifold harmonic analysis. 
  Left-hand side: the simulated electrostatic potential function on the protein surface. 
  Right-hand side: the linear combination of its Laplace-Beltrami eigenfunctions \rev{($\phi_1, \ldots, \phi_{10}, \ldots$)} with corresponding coefficients \rev{($c_{\phi_{1}}, \ldots, c_{\phi_{10}}, \ldots$)}.  
  \rev{Only a few selected terms are explicitly shown from the infinite sum. 
  Note that these eigenfunctions exhibit different spatial frequencies (resolutions).}
  }
\label{fig:fig2}
\end{figure}

Before moving on, we first explain some basic concepts behind our proposed representation learning framework. 
In \autoref{sec:shape_DNA}, we introduce molecular ``Shape-DNA''
and a set of basis functions inherent to the surface manifold.
In \autoref{sec:harmonic_analysis}, we illustrate how to apply harmonic analysis to decompose a function 
(defined on the molecular surface) into the linear combination of its ``Shape-DNA'' basis functions, as shown in \autoref{fig:fig2}.
These concepts enable us to represent a molecular surface as a 2D Riemannian manifold with associated functions,
which form the cornerstone of our proposed framework.

\vspace{-5pt}
\subsection{The Shape-DNA}\label{sec:shape_DNA}

The molecular surface 
\footnote{For instance, an isosurface of its electron density field, or the solvent-accessible/exclusive surface, etc.} 
can be viewed as a 2D Riemannian manifold ($\mathcal{M}$), which adopts a discrete set of eigenfunctions ($\phi$) that solves
\begin{equation}
\Delta \phi_i = \lambda_i \phi_i, \quad i = 0, 1, \ldots
\end{equation}
Here, $\Delta$ is the Laplace-Beltrami (LB) operator acting on surface scalar fields, 
defined as $\Delta f = -\mathrm{div} (\nabla f)$.
$\phi_0, \phi_1,\dots$ are a set of orthonormal eigenfunctions 
(i.e., $\langle \phi_i, \phi_j \rangle_\mathcal{M} = \delta_{ij}$).
And $0=\lambda_0 \leq \lambda_1 \leq \dots$ are the corresponding eigenvalues. 
%
These set of eigenvalues $\{\lambda_i\}$ of the LB operator are called the ``Shape-DNA'' and their corresponding eigenfunctions $\{\phi_i\}$ are unique to each shape.
\footnote{Two shapes may share the same LB eigendecomposition (isometries).
The uniqueness of eigendecomposition up to isometries 
is associated with the question ``Can One Hear the Shape of a Drum?''~\citep{kac1966can}}.
\rev{In other words, different molecules adopt different LB eigenfunctions.
Notably, two different 3D conformations of the same molecule may not have the same LB eigenfunctions, 
but a pair of chiral molecules (i.e., mirror image of each other) share the same LB spectrum.}

\autoref{fig:fig2} shows a few selected eigenfunctions ($\phi_1$, $\phi_{10}$, $\phi_{100}$) of a protein surface manifold on the right-hand side.
These eigenfunctions are intrinsic properties of the surface manifold, which remain invariant under rigid transformations to the molecule.
The eigenvalues $\lambda_i$ reflect the surface Dirichlet energy (defined as $\langle \Delta f, f\rangle_{ \mathcal{M}}$), 
which measures the smoothness of eigenfunction $\phi_i$ over $\mathcal{M}$.
The eigenvalues increase linearly, whose slope is roughly inversely proportional to the surface area (known as Weyl's asymptotic law). 
\rev{See \autoref{ap:shape_DNA} for more details about the ``Shape-DNA''.}

\subsection{Basics of Manifold Harmonic Analysis} \label{sec:harmonic_analysis}

Now, we introduce basic manifold harmonic analysis, a merit of using the Riemannian manifold representation.
``Harmonic analysis'' refers to the representation of functions as the superposition of some basic waves.
Specifically in our case, given the molecular surface manifold $\mathcal{M}$ and its LB eigenfunctions $\{\phi_i\}_{i=0}^\infty$,
any scalar-valued function $f$ that is square-integrable on $\mathcal{M}$ can be decomposed into a generalized Fourier series:
\begin{equation}
f(x) = \sum_{i=0}^{\infty} \langle f, \phi_i \rangle_\mathcal{M} \phi_i(x) \,.
\end{equation}
In other words, $f$ can be represented as the linear combination of the LB eigenfunctions.
The linear combination coefficient $\langle f, \phi_i \rangle_\mathcal{M}$ 
can be interpreted as the ``projection'' of $f$ onto the eigenfunction $\phi_i$, 
which reflects the contribution of this particular eigenfunction to synthesizing function $f$.

Interestingly, it is easy to notice that these eigenfunctions display different spatial frequencies (\autoref{fig:fig2}), 
i.e., $\phi_1$ varies slowly over the surface, while $\phi_{100}$ oscillates at a much higher frequency. 
This is analogous to the set of $\sin(kx)$ functions in the 1D case, 
where high frequency waves with larger $k$ values exhibit more oscillations within the period of length $2\pi$.
Hereafter we refer to LB eigenfunctions with (relatively) small/large eigenvalues as the low/high frequency components.

Therefore, by manipulating the linear combination coefficients, 
we could control the contribution of different frequency components to 
synthesizing function $f$.
For instance, with only low frequency components, the synthesized function will have a lower spatial resolution on the surface 
(i.e., more smoothed out, \autoref{fig:fig1} right panel, \rev{see \autoref{ap:harmonic_analysis} for technical details.}), and vice versa.
Synthesizing a new function $f$ with some selected frequency components is commonly known as wave filtering in signal processing,
which will be applied in our representation learning framework to realize multi-resolution encoding of the molecular surface.
We refer the readers to some excellent review papers for more details about geometry processing~\citep{Br:2017geometric,Ro:1997laplacian}.

\section{Methodology}

We realize that the aforementioned geometry processing methods are suitable for molecular surface representation.
However, a clear distinction between the 3D objects used in geometry processing research and the molecular systems 
is that molecules are not simply shapes -- 
their underlying atomic structures beneath the surface govern the molecular functionality.
In other words, both geometry (i.e., shape) and chemistry have to be considered for molecular encoding.
It has been shown that the molecular surface displays chemical and geometric patterns 
which fingerprint a protein's mode of interactions with other biomolecules~\citep{Ga:2020deciphering}.
Therefore, we formulate both geometric and chemical properties of a molecule as functions distributed on its surface manifold.

Then comes the question: how do we properly learn these geometric and chemical features on the molecular surface?
One viable solution is to emulate the message passing framework commonly used in GNNs, 
whose goal is to propagate information between distant surface regions to encode surface features at different scales.
To that end, we present two methods under the \model framework.
In \autoref{sec:HMP}, we introduce manifold harmonic message passing,
which makes use of harmonic analysis techniques to allow 
efficient multi-range communication between regions on the molecular surface regardless of its size. 
This enables \model to encode information within a molecule and be applied to molecule-level prediction tasks.
In \autoref{sec:method_docking}, we propose a \model-powered rigid protein docking pipeline.
We use this docking challenge to demonstrate the potential of our proposed representation learning method 
for modeling interactions between large biomolecules.

\subsection{Learning Harmonic Molecular Representations}\label{sec:HMP}

Given the 3D atomic structure of a molecule, \model returns 
(1) a discretized molecular surface manifold (i.e., a triangle mesh) with $N$ vertices 
\{$\mathbf{x}_1,\ldots,\mathbf{x}_N$\} $\subset \mathbb{R}^3$ 
and the corresponding faces; 
(2) a set of $n$ per-vertex features, 
$\mathsf{F} \in \mathbb{R}^{N \times n}$, 
which can be viewed as $n$ learned surface functions.
\rev{The discretized surface manifold as well as these features represent the geometry and chemistry of the underlying molecule,} 
and can be used for various downstream prediction tasks.

\rev{\textbf{Surface Preparation}}\quad
We use \texttt{MSMS}~\citep{ewing2010msms} to compute the molecular solvent-excluded surface as a triangle mesh with $N$ vertices.
Then, we compute the first $k$ LB eigenfunctions $\{\phi_i\}_{i\geq0}^{k-1}$ with ascending eigenvalues 
as described in~\cite{reuter2009discrete},
and stack them into an array $\Phi \in \mathbb{R}^{N \times k}$, where each column stores an eigenfunction.

Geometric features can be readily calculated given the surface mesh.
We compute the per-vertex mean curvature, Gaussian curvature, and the Heat Kernel Signatures as described in~\cite{sun2009concise}.
Local chemical environment is captured using a simple multilayer perceptron (MLP).
For each vertex, we encode its neighboring atoms within a predefined radius (e.g., 6\,\AA) through MLP,
then sum over the neighbors to obtain its chemical embedding. 
We use another MLP to combine the per-vertex initial features 
$\mathsf{F}^{\mathrm{inp}} \leftarrow \mathrm{MLP} (\mathrm{concat}(\mathsf{F}^{\mathrm{geom}}, \mathsf{F}^{\mathrm{chem}}))$,
$\mathsf{F}^{\mathrm{inp}} \in \mathbb{R}^{N \times n}$.
These $n$ features reflect the local geometric and chemical environment of each surface vertex, 
which will be used as input to the harmonic message passing module.
See more implementation details in \autoref{ap:surface_prep}.

The output of the surface preparation module includes (1) the molecular surface triangle mesh, 
(2) the surface Laplace-Beltrami eigenfunctions $\Phi$, and (3) the per-vertex features $\mathsf{F}^{\mathrm{inp}}$.

\begin{figure}[tb]
  \centering
  \includegraphics[width=0.9\linewidth]{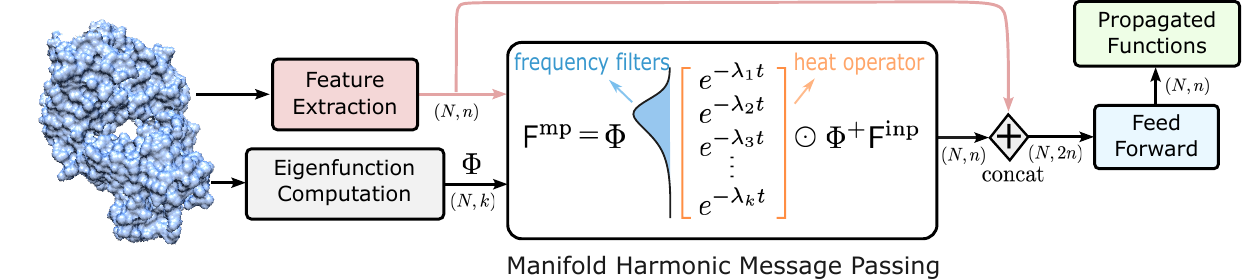}
  \caption{
  \rev{\model workflow. Given a molecular surface mesh with $N$ vertices, 
  we compute the first $k$ Laplace-Beltrami eigenfunctions 
  (column-wise stacked into an array $\Phi$, and $\Phi^{+}$ denotes its Moore-Penrose pseudo-inverse, 
  see \autoref{ap:harmonic_analysis} for discrete calculations) 
  with ascending eigenvalues,
  and extract $n$ initial surface features $\mathsf{F}^{\mathrm{inp}}$ through MLP.
  Then, we apply neural network-learned spectral filters to propagate the features over the surface to achieve message passing. 
  Note that each feature channel has a unique Gaussian frequency filter and propagation time $t$.
  Relevant tensor sizes are indicated in parentheses. Multiple message passing blocks can be stacked for better representations.}
  }
  \label{fig:fig3}
  \vspace{-8pt}
\end{figure}

\textbf{Harmonic Message Passing}\quad
Our proposed harmonic message passing mechanism is closely related to the heat diffusion process on an arbitrary surface.
Joseph Fourier developed spectral analysis methods to solve the heat equation $\partial f/\partial t + \Delta f = 0$, 
where $f$ is some heat distributed on the surface.
This concise partial differential equation describes how a heat distribution $f$ evolves over time,
whose solution can be expressed using the heat operator $\exp(-\Delta t)$, 
i.e., $f(t) = \exp(-\Delta t) f_0$ for initial heat distribution $f_0$ at $t=0$.
Intuitively, heat will flow from hot regions to cool regions on the surface. 
As time approaches infinity, the heat distribution $f$ will converge to a constant value (i.e., the global average temperature on the surface), 
assuming that total energy is conserved.

In fact, heat diffusion can be thought of as a message passing process, 
where surface regions with different temperatures communicate with each other and propagate the initial heat distribution deterministically.
The heat exchange rate is dependent on the difference in temperature (determined by the LB operator), 
while the message passing distance is determined by the heat diffusion time $t$.
Following this idea, we generalize the heat diffusion process by proposing a function propagation operator 
$\mathcal{P}$ with neural network-learned frequency filter $F_{\theta}(\lambda)$:
\begin{equation}
\small
    \mathcal{P} f = \sum_i F_{\theta}(\lambda_i) \langle f, \phi_i \rangle_{\mathcal{M}} \phi_i \,,
    \label{eq:prop_operator}
\end{equation}
\begin{equation}
\small
    F_\theta(\lambda) = \exp\left(-\frac{(\lambda - \mu)^2}{\sigma^2}\right) \cdot \exp(-\lambda t) \qquad \mathrm{where} \quad \theta = (\mu, \sigma, t)\,.
    \label{eq:freq_filter}
\end{equation}
As shown in \autoref{eq:prop_operator}, the input function $f$ is first 
expanded as the linear combination of the LB eigenfunctions with coefficients $\langle f, \phi_i \rangle_{\mathcal{M}}$.
We then learn a spectral-space frequency filter $F_\theta(\lambda)$, which is a function of the corresponding eigenvalue $\lambda$.
\rev{Here we abuse the usage of frequency, which actually refers to the LB eigenvalues.}
The filter $F_\theta(\lambda)$ consists of two components (\autoref{eq:freq_filter}): 
a Gaussian frequency filter (with parameters $\mu$ and $\sigma$), and the heat operator part $e^{-\lambda t}$ (with parameter $t$).
The matrix representation of the function propagation operator $\mathcal{P}$ is plotted in \autoref{fig:fig3}.

\begin{wrapfigure}{R}{0.4\textwidth}
\includegraphics[width = 0.4\textwidth]{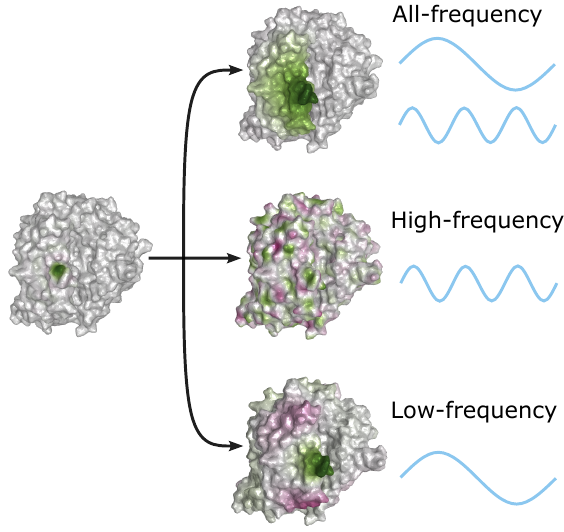}
  \caption{Examples of versatile message passing outcomes using different frequency filter settings. 
  }
\label{fig:fig4}
\vspace{-8pt}
\end{wrapfigure}

For each input function $f$ within $\mathsf{F}^{\mathrm{inp}}$, 
the neural network learns a unique set of parameters ($\mu$, $\sigma$, $t$) through backpropagation.
In \autoref{fig:fig4}, we showcase a few examples of surface functions obtained 
by applying different frequency filters to an initial delta function.

The Gaussian frequency filter allows the network to propagate the input function 
along some selected eigenfunctions with eigenvalues close to $\mu$, 
while the number of selected eigenfunctions is determined by the filter width $\sigma$.
Owing to the multi-resolution nature of the LB eigenfunctions with different spatial frequencies, 
this filter will help capture surface functions at different resolutions.

The heat operator part governs the communication distance.
With longer propagation time, function $f$ will be more averaged out towards the global mean, leading to a smoothed function.
In addition, the heat operator is by definition a low-pass filter, where components with larger eigenvalues decay faster.
This makes eigenfunctions with large eigenvalues contribute less significantly during message passing.
Therefore, the combination of the Gaussian frequency filter and heat operator could help the network focus on 
some higher frequency components~\citep{aubry2011wave}.

As demonstrated, \model is able to represent a variety of surface functions through harmonic message passing.
The output of this module has the same size as the input, $\mathsf{F}^{\mathrm{mp}} \in \mathbb{R}^{N \times n}$, 
where each channel respectively contains the propagated version of its input function.
These features represent the neighboring geometric and chemical environment across multiple spatial scales, 
and can be used for surface property-related tasks, e.g., protein binding site prediction (see \autoref{sec:method_docking}).
In addition, molecule-level representations could be obtained through global pooling (see \autoref{sec:exp_qm9}).

\subsection{Learning Surface Correspondence for Rigid Protein Docking}\label{sec:method_docking}

\begin{figure}[t]
  \centering
  \includegraphics[width=0.98\linewidth]{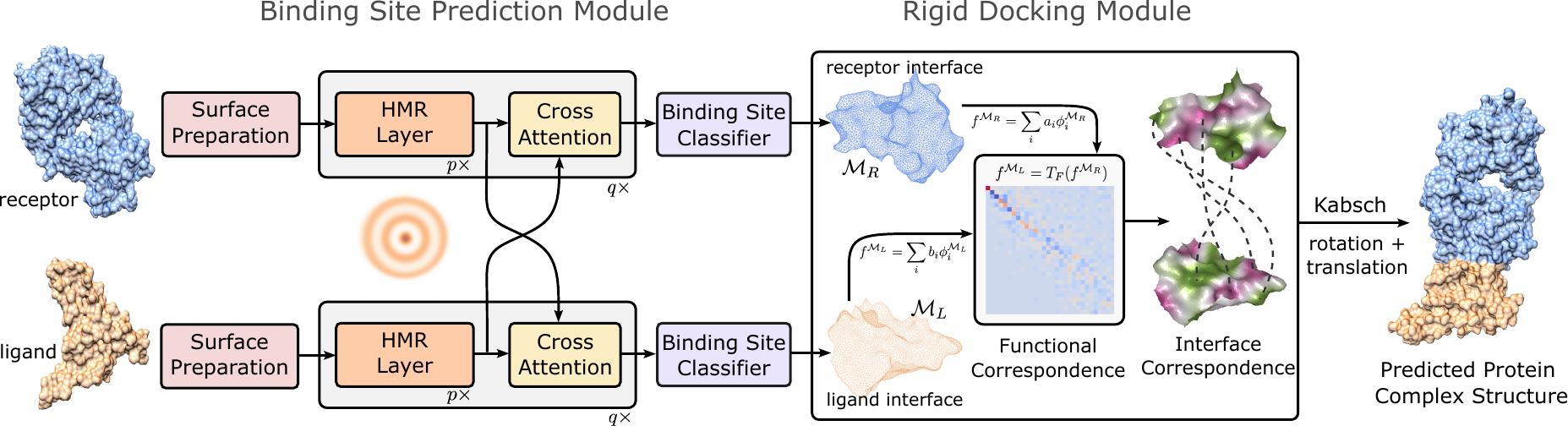}
  \caption{
  The rigid protein docking pipeline. Given the surface of the ligand and receptor proteins, 
  we apply multiple \model and cross attention layers to allow communication within and between the surfaces. 
  The learned surface representations are then used to predict the binding interfaces 
  and establish functional correspondence using functional maps. 
  Rigid docking is achieved by converting the functional correspondence to rigid transformations which aligns the predicted interfaces.
  }
  \vspace{-6pt}
  \label{fig:fig5}
\end{figure}

In this section, we demonstrate how to learn surface correspondence for molecular matching.
Specifically, we introduce a surface-based rigid protein docking workflow (\autoref{fig:fig5})
powered by \model. 
\rev{Rigid protein docking is a significant problem in structural biology, 
whose goal is to predict the pose of the protein complex based on the structure of the ligand and receptor proteins.}
It has been shown that protein-protein interfaces exhibit similar geometric and chemical patterns~\citep{Ga:2020deciphering}.
In other words, two proteins may interact if part of their surfaces display similar shapes and chemical functions (i.e., functional correspondence).
\rev{This is similar to solving a puzzle problem, where both shape and pattern of the missing piece have to match in order to complete the puzzle.}

Following this idea, we propose to realize rigid protein docking in two consecutive steps:
(1) given two protein surfaces, predict the region where binding might occur (i.e., binding site prediction, \rev{locating the missing piece});
(2) establish functional correspondence between the ligand/receptor binding surfaces, 
and convert it to real-space vertex-to-vertex correspondence (\rev{shape/pattern matching}).
Rigid docking could then be achieved by aligning the corresponding binding site surface vertices.

\textbf{Binding Site Prediction}\quad
Given the ligand and receptor protein surface meshes, we first predict the regions where they interact, which is a per-vertex binary classification problem.
We iteratively apply \model and cross-attention layers~(\autoref{fig:fig5}) to encode the surfaces with intra- and inter-surface communications.
Next, we use the learned features on each vertex to classify whether it belongs to the binding interface. 
Detailed descriptions of this module are available in \autoref{sec:details_of_BSP}.

The output of the binding site prediction module includes:
a) the Receptor~(Ligand) surface mesh of the binding interface, which is a submanifold of the entire protein surface, represented as
$\mathcal{M}_R$~($\mathcal{M}_L$) with $N_R$~($N_L$) vertices, and
b) neural network-learned features or so-called surface functions 
$\mathsf{F}^{\mathcal{M}_R}$ and $\mathsf{F}^{\mathcal{M}_L}$ distributed on the receptor and ligand binding interfaces, respectively.

\textbf{Rigid Docking with Functional Maps}\quad
We know that protein interfaces $\mathcal{M}_R$ and $\mathcal{M}_L$ exhibit similar geometry 
and also adopt a set of $n$ corresponding functions 
$\mathsf{F}^{\mathcal{M}_R} \in \mathbb{R}^{N_R \times n}$ and 
$\mathsf{F}^{\mathcal{M}_L} \in \mathbb{R}^{N_L \times n}$ (\autoref{fig:fig5}).
Intuitively, if we could somehow align these set of corresponding functions, 
then we have found a way to align the protein binding interfaces.
To that end, we employ functional maps to ``align'' $\mathsf{F}^{\mathcal{M}_R}$ and $\mathsf{F}^{\mathcal{M}_L}$ in spectral domain.

Specifically, given the truncated LB eigenfunctions $\Phi^{\mathcal{M}_R} \in \mathbb{R}^{N_R \times k}$ and 
$\Phi^{\mathcal{M}_L} \in \mathbb{R}^{N_L \times k}$ of the receptor and ligand interface manifolds,
we respectively compute the spectral representation of learned functions as
$\mathsf{A} = (\Phi^{\mathcal{M}_R})^{+}\mathsf{F}^{\mathcal{M}_R}$,
and $\mathsf{B} = (\Phi^{\mathcal{M}_L})^{+}\mathsf{F}^{\mathcal{M}_L}$,
where $+$ denotes the Moore-Penrose pseudo-inverse, and $\mathsf{A}, \mathsf{B} \in \mathbb{R}^{k \times n}$.
Functional correspondence ($\mathsf{C}$) can be obtained by minimizing:
\[
\mathsf{C} = \argmin_{\mathsf{C} \in \mathbb{R}^{k \times k}} \| \mathsf{C}\mathsf{A} - \mathsf{B} \|_\mathrm{F} \,,
\]
where $\|\cdot\|_\mathrm{F}$ denotes the Frobenius norm.
Once the functional mapping (i.e., the $\mathsf{C}$ matrix) is recovered through numerical optimization, 
vertex-to-vertex correspondence between the receptor and ligand surfaces can be established
by mapping indicator functions of vertices on $\mathcal{M}_R$ to those on $\mathcal{M}_L$. 
In practice, we adopted a slightly more complicated functional maps approach, which is illustrated in \autoref{ap:fmaps}.
Finally, we perform rigid docking by aligning the proteins according to the vertex-to-vertex surface correspondence using the Kabsch algorithm~\citep{Ka:1976solution}.

\section{Experiments}

\subsection{QM9 Molecular Property Regression}\label{sec:exp_qm9}

\begin{wraptable}{r}{6.1cm}
\vspace{-40pt}
\centering
\caption{QM9 Mean Absolute Error}\label{tab:qm9}
\footnotesize
\setlength{\tabcolsep}{2pt}
\begin{tabular}{lccccc}
\toprule  
Task & $\alpha$ & $\Delta\varepsilon$ & $\varepsilon_\mathrm{HOMO}$ & $\mu$ & $C_v$\\
Unit & bohr$^3$ & meV & meV & D & cal/(mol K) \\
\midrule
SchNet      & .235 & 63   & 41   & .033  & .033 \\
NMP         & .092 & 69   & 43   & .030  & .040 \\
Cormorant   & .085 & 61   & 34   & .038  & .026 \\
SE(3)-Tr.   & .142 & 53   & 35   & .051  & .054\\
SEGNN       & .060 & 42   & 24   & .023  & .031 \\ 
\midrule
\rev{\textbf{\model}} & .102 & 59 & 37 & .037  & .040\\  
\bottomrule
\end{tabular}
\vspace{-20pt}
\end{wraptable}

We employ \model to perform property regression tasks on the QM9 dataset~\citep{Ra:2014quantum}.
We use the same data split and atomic features as ~\cite{Sa:2021n}.
We compare our results with both invariant and equivarient networks as shown in \autoref{tab:qm9},
including SchNet~\citep{schutt2018schnet}, NMP~\citep{Gi:2017neural}, Cormorant~\citep{An:2019cormorant},
SE(3)-Transformer~\citep{Fu:2020se}, 
and SEGNN~\citep{brandstetter2021geometric}.
See \autoref{ap:qm9} for the complete table and experimental setup.
Interestingly, despite \model completely discards the bonding information 
and only performs massage passing over the molecular surface, 
it still shows comparable performance in predicting these molecular properties.

\vspace{-20pt}
\rev{\subsection{Ligand-binding Pocket Classification}}
\begin{wrapfigure}{R}{0.4\textwidth}
\vspace{-25pt}
\includegraphics[width = 0.4\textwidth]{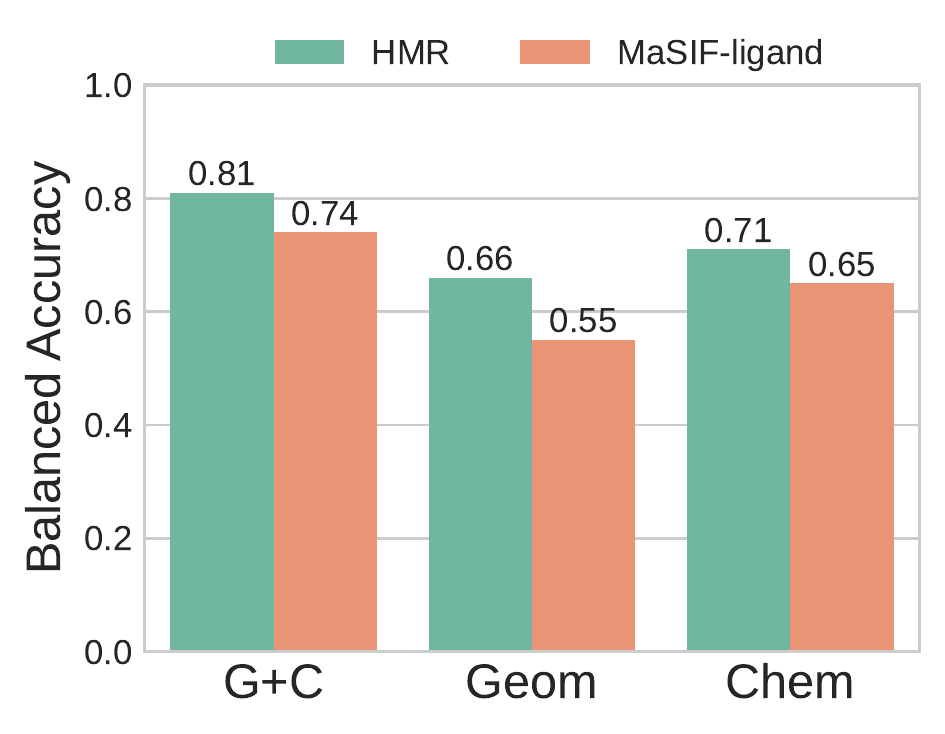}
  \caption{Balanced accuracy of ligand-binding protein pocket classification. 
  Models with different input features are compared: ``G+C'' includes both geometric and chemical features, 
  while ``Geom''/``Chem'' only contains geometric/chemical features.}
  \vspace{-10pt}
\label{fig:fig6}
\end{wrapfigure}

Next, we showcase the representation power of \model in predicting the ligand-binding preference of protein pockets. 
The preference of a protein binding to a specific small molecule (ligand) is associated with the geometric and chemical environment at the binding region (i.e., binding pocket).
As defined in \cite{Ga:2020deciphering}, we classify a protein pocket as one of seven ligand-binding types 
(ADP, COA, FAD, HEM, NAD, NAP, and SAM) using its surface information.
The dataset contains 1,634 training, 202 validation, and 418 test cases. 
We compare the performance of \model to MaSIF-ligand~\citep{Ga:2020deciphering},
which is a geodesic convolutional neural network-based model.
A set of features similar to MaSIF-ligand are used in our model:
hydrophobicity score and partial charge as chemical features; 
mean/Gaussian curvature and the Heat Kernel Signatures as geometric features.

As shown in \autoref{fig:fig6}, the balanced accuracy of our model is consistently better than that reported in \cite{Ga:2020deciphering}, 
suggesting that the proposed \model framework can more effectively encode protein surface information through harmonic message passing 
over the molecular surface Riemannian manifold. In addition,
we draw a similar conclusion that both geometric and chemical information of the binding pockets 
are important in predicting the type of its binding molecules.
See \autoref{ap:masif-ligand} for implementation details and more result analysis.

\subsection{Rigid Protein Docking}\label{sec:res_docking}

Finally, we evaluate \model on a more challenging task: rigid protein docking.

\textbf{Experimental setup}\quad
\model is trained on a modified version of Database of Interacting Protein Structures (DIPS)~\citep{townshend_dips_2019}
and evaluated on the gold-standard Docking Benchmark 5.5 (DB5.5)~\citep{guest_expanded_db55_2021}.
We compare our model with the state-of-the-art GNN-based deep learning model \textsc{EquiDock}~\citep{Ga:2021independent} 
and two traditional docking methods, \textsc{Attract}~\citep{de_vries_attract_2015} and \textsc{Hdock}~\citep{yan_hdock_2020}. 
To evaluate docking performance, we compute the Complex and Interface root-mean-square deviation (RMSD) following~\cite{Ga:2021independent}, 
and calculate DockQ following~\cite{basu_dockq_2016}.
We also report the success rate indicating whether the result achieves ``Acceptable'' or higher according to~\cite{lensink_docking_capri_2013}.

\begin{table}[b]
\vspace{-6pt}
\setlength{\tabcolsep}{3.5pt}
\footnotesize
\caption{Rigid Prediction Results on Docking Benchmark 5.5}
\label{table:benchmark}
\begin{center}
\vspace{-6pt}
\begin{tabular}{lcccccccc}
\toprule
{}                & \multicolumn{2}{c}{Complex RMSD $\downarrow$} & \multicolumn{2}{c}{Interface RMSD $\downarrow$} & \multicolumn{2}{c}{DockQ $\uparrow$}   & Success rate $\uparrow$ & \rev{Inference}\\ \cmidrule(lr){2-3} \cmidrule(lr){4-5} \cmidrule(lr){6-7}
{Model}                 &       Median &   Mean &         Median &   Mean &     Median &  Mean       & ($\geq$ Acceptable)  & \rev{time (sec)}  \\
\midrule
\textbf{\model (Top 1)} &        12.01 &  12.09 &          10.90 &  11.43 &    0.06 &  0.22 &                          0.28 & 5.8 \\
\textbf{\model (Top 3)} &         8.44 &   9.68 &           7.50 &   8.70 &    0.10 &  0.26 &                          0.33 & 6.6\\
\textsc{EquiDock}*       &        15.90 &  17.18 &          14.45 &  14.85 &    0.02 &  0.04 &                           0.00 & 3.9\\
\textsc{Attract}        &         8.99 &  11.11 &          10.31 &  11.67 &    0.07 &  0.42 &                           0.44 & 882.7\\
\textsc{Hdock}          &         0.39 &   5.97 &           0.32 &   5.68 &    0.97 &  0.71 &                          0.73 & 884.9\\
\bottomrule
\end{tabular}
\end{center}
\quad\quad\quad\quad *\textsc{EquiDock} model is trained on DIPS, provided by the authors
\end{table}

\begin{figure}[t]
  \centering
 \includegraphics[width = 1.0\textwidth]{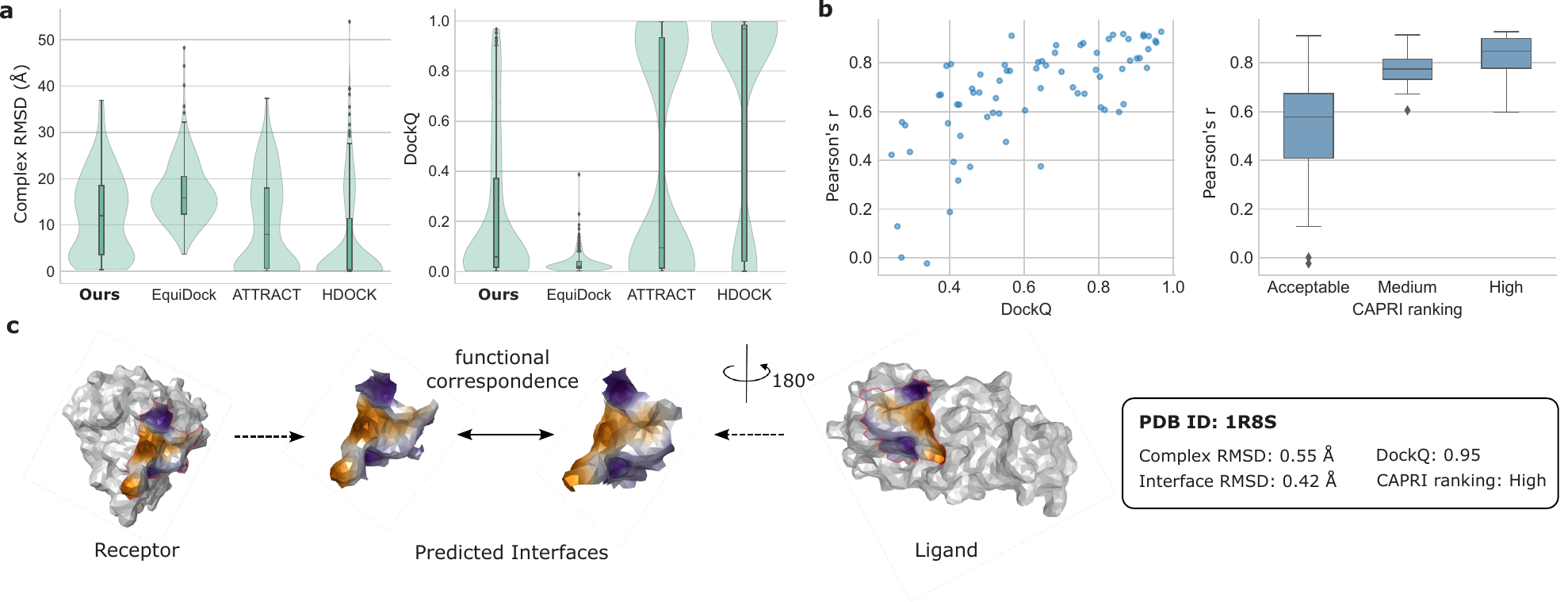}
  \caption{Rigid docking performance analysis. 
  \textbf{a}. Distribution of Complex RMSD and DockQ scores for poses predicted using 
  \model (ours, Top 1), \textsc{EquiDock}, \textsc{Hdock}, and \textsc{Attract}. 
  \textbf{b}. Correlations between learned functions on the predicted ligand-receptor interfaces. 
  For each test case, we calculate the correlation as the Pearson's $r$ between 
  the learned function values on ligand interface vertices and 
  their nearest receptor vertices (corresponding vertices), 
  averaged over 128 hidden channels. 
  \textbf{c}. A showcase of functional correspondence between interfaces for a well-docked case.}
\label{fig:fig7}
\vspace{-10pt}
\end{figure}

\textbf{Results}\quad
Model performance are summarized in \autoref{table:benchmark} and \autoref{fig:fig7}a. 
\model (Top 1) outperforms GNN-based \textsc{EquiDock} model under all metrics. 
Notably, \model achieves a much higher success rate, indicating more test cases have results close to the ground truth complex structure. 
We note that traditional methods still exceed in terms of docking performance but at greater computational cost (\autoref{tab:docking_time}).
Further experiments show that the harmonic message passing mechanism learns to propagate information at different scales (\autoref{fig:figa3}) 
and is critical to the effectiveness of the model (\autoref{tab:mp-mech}). 
Our feature ablation studies show that 
chemical information is particularly important in the rigid protein docking task (\autoref{tab:feat-ablation}). 
Compared to \textsc{EquiDock}, the novel framework of \model and optimized training dataset collaboratively contribute to the higher performance (\autoref{tab:cross-exp}).

We further examine the learned features at the predicted binding interfaces. 
As shown in \autoref{fig:fig7}bc, surface functions on the two binding sites are highly correlated, 
confirming the good alignment achieved using functional maps. 
Cases with higher docking quality 
show stronger interface feature correlations, 
suggesting our \model is able to learn complex surface interaction patterns, 
which also supports the claims in \cite{Ga:2020deciphering} about the significant role of protein surface.

\model predicts multiple binding sites for some proteins 
\rev{(either due to certain protein symmetries or model uncertainty)}. 
Therefore, we also assess the model performance by including candidate poses from top 3 binding site pairs, 
ranked by the mean probability predicted by the binding site classifer. 
The best scores from top 3 poses show that the performance of \model is competitive to \textsc{Attract}. 
See \autoref{ap:docking} for more detailed analysis on the rigid protein docking experiment.

\vspace{-6pt}
\section{Conclusions and Outlook}
\vspace{-6pt}

We presented \model, a powerful surface manifold-based molecular representation learning framework.
By integrating geometric and chemical properties as functions distributed on the molecular surface manifold, 
and applying harmonic message passing in spectral domain, 
we achieve multi-resolution molecular representations.
\model shows promising performance in molecular property prediction, protein pocket classification, and molecular matching tasks.
Our work highlights an important aspect of molecular ``structure-activity'' relationship, that is -- ``shape-activity'' relationship.
We think this is particularly significant for large biomolecules, where the surface shape and chemical patterns determine 
some fundamental biological activities, such as protein-protein interactions.

\rev{
The \model framework could serve as a complementary method to GNN/NMP-based models in solving challenges for complex biological systems, 
which exhibits its unique advantages and shortcomings.
One of the foreseeable challenges in further developing the \model framework is
the efficient computation of the ``Shape-DNA'' (i.e., solving a second-order partial differential equation)
in order to incorporate protein dynamics.
Since either a change of the protein backbone or some side chains near the surface may alter the entire ``Shape-DNA'', 
which needs to be recomputed upon protein conformational change.
This is particularly important for large biomolecules where the surface shape undergoes significant changes 
(e.g., the Complementarity-Determining Regions of an antibody upon binding with an antigen, or the allosteric site of some enzymes).
To that end, we call for more research attention to surface manifold-based molecular representation learning.
}

\newpage

\section*{Acknowledgements}
The authors thank Dr. Hang Li and Dr. Quanquan Gu for their insightful comments. Hao Zhou is supported by Vanke Special Fund for Public Health and Health Discipline Development, Tsinghua University (NO.20221080053), Guoqiang Research Institute General Project, Tsinghua University (No. 2021GQG1012).

\section*{Reproducibility Statement}

The code and data are available at \href{https://github.com/GeomMolDesign/HMR}{https://github.com/GeomMolDesign/HMR}.
QM9 dataset is provided at ~\url{https://springernature.figshare.com/ndownloader/files/3195389}.
The dataset for the ligand-binding pocket classification is provided at ~\url{https://zenodo.org/record/2625420} and the split used by MaSIF is at \url{https://github.com/LPDI-EPFL/masif/tree/master/data/masif_ligand/lists}.
DIPS dataset can be downloaded from the following website~\url{https://github.com/BioinfoMachineLearning/DIPS-Plus}. \textsc{EquiDock} model and checkpoints can be downloaded from ~\url{https://github.com/octavian-ganea/equidock_public}. \textsc{Attract} can be downloaded from~\url{https://github.com/sjdv1982/attract}. \textsc{Hdock} is implemented using its local version \texttt{HDOCKlite}, which can be downloaded from ~\url{http://huanglab.phys.hust.edu.cn/software/hdocklite/}. DockQ can be downloaded from~\url{https://github.com/bjornwallner/DockQ/}. 

\bibliography{iclr2023_conference}
\bibliographystyle{iclr2023_conference}

\newpage
\appendix

\setcounter{table}{0}
\renewcommand{\thetable}{\Alph{section}.\arabic{table}}
\setcounter{figure}{0}
\renewcommand{\thefigure}{\Alph{section}.\arabic{figure}}
\setcounter{equation}{0}
\renewcommand{\theequation}{\Alph{section}.\arabic{equation}}

\begin{center}
    \rev{\Large{\textbf{Appendix}}}
\end{center}

\section{The Molecular Shape-DNA}\label{ap:shape_DNA}

\textbf{Riemannian Manifold}\quad
A manifold is a space that is locally flat but not necessarily globally flat. 
Formally speaking, a $d$-dimensional manifold $\mathcal{M}$ is a topological space 
where each point $p\in\mathcal{M}$ has a neighborhood that is homeomorphic to a $d$-dimensional Euclidean space~\citep{lee2013smooth}, which is equivalent to the tangent space at $p$ and is denoted by $T_p\mathcal{M}$.

We can further assign a positive definite inner product $g:T_p\mathcal{M} \times T_p\mathcal{M} \to \mathbb{R}$ on every tangent space, and this inner product is called a Riemannian metric. A manifold equipped with a Riemannian metric is called a Riemannian manifold. 
Intuitively, the Riemannian metric provides a measurement of the velocity when a particle moves on the manifold, and many other
quantities can therefore be defined. For example, for any tangent vector $X_p\in T_p\mathcal{M}$, the quantity $|X_p| := \sqrt{g(X_p,X_p)}$ can be interpreted as the traveling speed of a particle when passing through $p$. Hence, the traveling distance along a curve (i.e., the length of a curve) can be defined as the integral of $|X_p|$ along the curve. On a Riemannian manifold, quantities that can be expressed in terms of the Riemannian metric are called intrinsic (e.g., geodesic distance). 

When a manifold is realized in the Euclidean space, a natural Riemannian metric can be induced from the ambient Euclidean space. We always refer to this induced metric when talking about a Riemannian manifold in the following appendices.

\textbf{The Lalpace-Beltrami Operator}\quad
We denote a real-valued scalar function on the manifold $\mathcal{M}$ by $f$. 
Given two functions $f_1, f_2$ on the manifold, we can define the inner product $\langle f_1, f_2 \rangle_{\mathcal{M}} = \int_\mathcal{M} f_1(x) f_2(x) \mathrm{d} \mu(x) $, where the area element $\mathrm{d} \mu$ is induced by the Riemannian metric. 
We denote by $L^2(\mathcal{M}) = \{f: \mathcal{M} \to \mathbb{R} \,|\, \langle f, f \rangle_\mathcal{M} < \infty \}$ the space of square-integrable functions on $\mathcal{M}$.
On a Riemannian manifold $\mathcal{M}$, we can generalize the usual Euclidean gradient $\nabla f$ and the positive semi-definite Laplace operator $\Delta f = -\mathrm{div}(\nabla f)$ to the intrinsic gradient $\nabla_\mathcal{M} f$ and the Laplace-Beltrami (LB) operator $\Delta_\mathcal{M} f$, respectively~\citep{petersen2006riemannian}. 

The LB operator admits a discrete set of eigenfunctions that solves
\begin{equation}
\Delta_\mathcal{M} \phi_i (x) = \lambda_i \phi_i(x) \qquad x \in \mathcal{M}
\end{equation}
with homogeneous Neumann boundary conditions if $\mathcal{M}$ has a boundary. 
Here, $0=\lambda_1 \leq \lambda_2 \leq \dots$ are eigenvalues and $\phi_1, \phi_2,\dots$ are the corresponding eigenfunctions. 
The LB eigenfunctions form an orthonormal basis of $L^2(\mathcal{M})$, i.e., $\langle \phi_i, \phi_j \rangle_\mathcal{M} = \delta_{ij}$.

\begin{figure}[b]
\centering
\vspace{-6pt}
\includegraphics[width=0.98\textwidth]{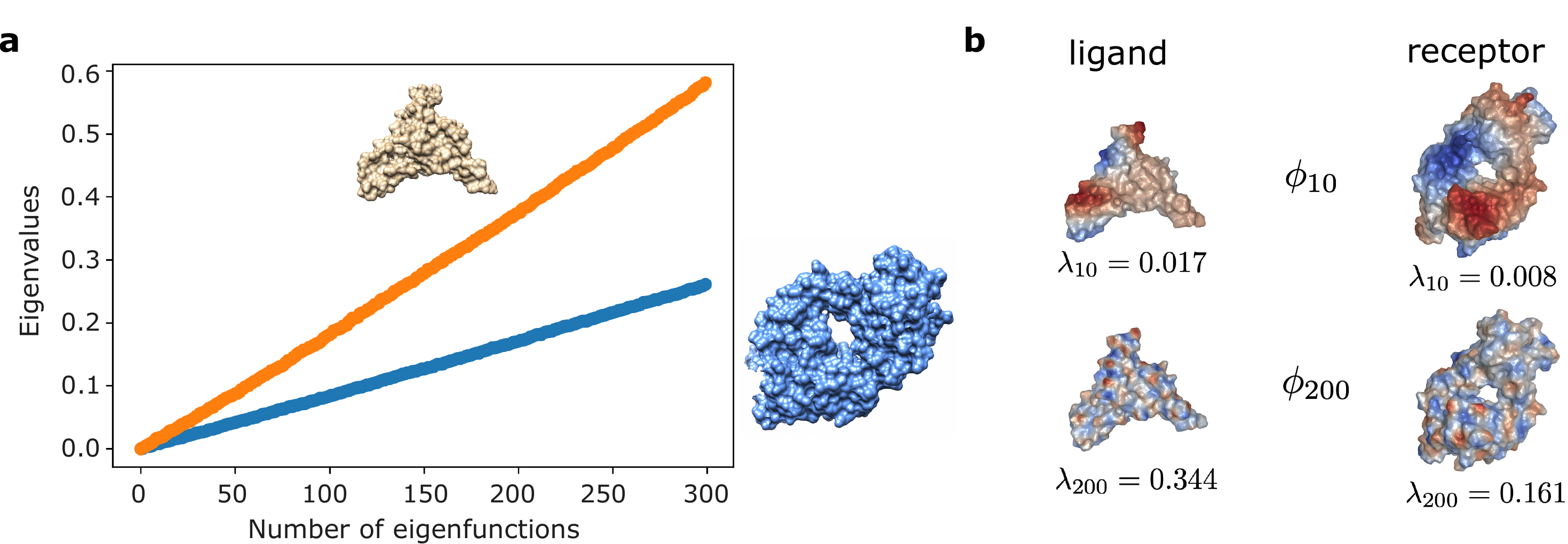}
  \caption{
  \textbf{a} Weyl's asymptotic law, where the protein with larger surface area exhibits a slower eigenvalue growth rate.
  \textbf{b} The 10$^{\mathrm{th}}$ and 200$^{\mathrm{th}}$ eigenfunction of the ligand and receptor surface manifold, respectively. 
  Eigenfunctions with similar eigenvalues exhibit similar spatial frequencies regardless of the surface area, 
  since they have similar smoothness in the sense of Dirichlet energy.
  }\label{fig:figA1}
\end{figure}

\textbf{Molecular Shape DNA}\quad
Intuitively, the LB eigenfunctions are the smoothest functions on the surface,
where smoothness is measured by the Dirichlet energy ($\langle \Delta f, f\rangle_{ \mathcal{M}}$).
Eigenfunctions with smaller eigenvalues are more smooth (i.e., with lower spatial frequency).
However, these eigenfunctions are not directly comparable across different manifolds.
\autoref{fig:figA1}a shows the first 300 eigenvalues of a pair of ligand and receptor protein surfaces (PDB ID: 3V6Z).
The protein with smaller surface area exhibits a larger slope, known as Weyl's asymptotic law.
\autoref{fig:figA1}b presents a few eigenfunctions with correspondending eigenvalues for the ligand and receptor molecules.
Since different eigenfunctions have their unique spatial resolutions, 
in our experiments we compute all eigenfunctions with eigenvalues smaller than a predefined value (determined empirically) 
to guarantee that molecules of different sizes have the same highest spatial resolution in their eigenfunctions.

\setcounter{table}{0}
\setcounter{figure}{0}
\setcounter{equation}{0}

\section{Implementing Manifold Harmonic Analysis}\label{ap:harmonic_analysis}

\textbf{Laplace-Beltrami Eigendecomposition}\quad 
Given a Riemannian manifold $\mathcal{M}$ and its Laplace-Beltrami operator $\Delta_\mathcal{M}$, the Laplacian eigenvalue problem we consider states as
$$\Delta_\mathcal{M} f = \lambda f ,$$
with homogeneous Neumann boundary condition.

To realize discrete calculations, we approximate the manifold with a triangle mesh consisting of $N$ vertices $\{\mathbf{x}_i\}_{i=1}^N \subset \mathcal{M}$, and the corresponding faces.
We then solve the discretized eigenvalue problem using a linear finite element method (FEM)~\citep{reuter2009discrete}.
The discretized equation we obtain is the following generalized eigenvalue problem
\begin{equation}\label{eqn:appB_eig}
\mathsf{A}_{\mathrm{cot}} \mathsf{f} = \lambda \mathsf{B} \mathsf{f},\quad \mathsf{f} := \left( f(\mathbf{x}_i) \right)^N_{i=1} \,,
\end{equation}
where $\mathsf{A}_{\mathrm{cot}}$ is the stiffness matrix with cotagent weights,
\begin{equation*}
    \mathsf{A}_{\mathrm{cot}}(i, j) :=
    \begin{cases}
    \frac{1}{2}(\cot \alpha_{ij} + \cot \beta_{ij}) & \text{if }(i, j) \text{ is an edge} \\
    \\
    -\sum_{k\in \mathcal{N}(i)} A_{\mathrm{cot}}(i, k) & \text{if }i = j\\
    \\
    0 & \text{otherwise} \,,
    \end{cases}
\end{equation*}
and $\mathsf{B}$ is an $N\times N$ sparse mass matrix which is associated with the weight of each surface vertex,
\begin{equation*}
    \mathsf{B}(i, j) :=
    \begin{cases}
    \frac{1}{12}(|t_1| + |t_2|) & \text{if } (i, j) \text{ is an edge} \\
    \\
    \frac{1}{6}\sum_{t_k\in T(i)} |t_k| & \text{if }i = j\\
    \\
    0 & \text{otherwise} \,.
    \end{cases}
\end{equation*}
Here $\alpha_{ij}$ and $\beta_{ij}$ are the two angles opposite to the edge $(i,j)$, and $\mathcal{N}(i)$ denotes the vertices that are adjacent to vertex $i$. The set $T(i)$ contains all the triangles that have $i$ as its vertex, and $|t_i|$ is the area of the triangle $t_i$. We also use $t_1$ and $t_2$ to denote the triangles that share the edge ($i$, $j$).

Such generalized symmetric eigenvalue problem can be solved with commonly used numerical simulaiton packages (e.g., \texttt{scipy}).
We can find non-negative eigenvalues $\mathsf{\Lambda}$ and eigenvectors $\mathsf{Z}$ such that
$$
\mathsf{Z}^{\top} \mathsf{A}_\mathrm{cot} \mathsf{Z} = \mathsf{\Lambda}, \quad \mathsf{Z}^{\top}\mathsf{B} \mathsf{Z} = \mathsf{I}.
$$
$\mathsf{I}$ is the identity matrix, $\mathsf{\Lambda} := \mathrm{diag}(\lambda_0,\lambda_1,\dotsc)$ is the diagonal matrix of the eigenvalues,
and the matrix $\mathsf{Z} \in \mathbb{R}^{N\times N}$ has the eigenvectors of~\autoref{eqn:appB_eig} as its column vectors.
Note that, unlike other conventional orthonormal basis (i.e., $\mathsf{Q}^{\top}\mathsf{Q} = \mathsf{I}$), here $\mathsf{Z}$ forms an orthonormal basis w.r.t. the mass matrix $\mathsf{B}$, that is, $\mathsf{Z}^{\top} \mathsf{B} \mathsf{Z} = \mathsf{I}$.

Practically, we do not need to store the entire $N \times N$ eigenvector matrix $\mathsf{Z}$. 
Just like in Fourier series expansion, a truncated Fourier basis with finite terms can be used to approximate the original signal. 
The number of basis that we keep determines the resolution of the synthesized signal, where typically high-frequency components are truncated.
In our case, we empirically determine the number of eigenvectors to keep for different molecular systems, which is also task-dependent.

\newpage
\textbf{Resolution Tuning with Harmonic Analysis}\quad 
We now explain how to realize surface resolution tuning under our representation framework (e.g., how to make \autoref{fig:fig1}).

Given a molecular surface triangle mesh, we first compute its Laplace-Beltrami eigendecomposition as described above.
Under the discrete setting, we refer to the eigenfunctions as eigenvectors.
We obtain a set of truncated Laplace-Beltrami eigenvectors $\mathsf{Z} \in \mathbb{R}^{N \times k}$ ($k$ is the number of eigenvectors we keep),
the associated $k$ eigenvalues $\{\lambda_i\}^{k-1}_{i=0}$, and the sparse mass matrix $\mathsf{B} \in \mathbb{R}^{N \times N}$.

Let $\mathsf{f} \in \mathbb{R}^{N}$ be the initial function of interest (stored as an $N$-dimensional array), which is distributed on the underlying molecular surface (e.g., electrostatic potential, a scalar value at each surface vertex). The spectral representation of function $\mathsf{f}$ can be calculated as
$$
\mathsf{f}^{\mathrm{spec}} = \mathsf{Z}^{\top} \mathsf{B} \mathsf{f},\quad \mathsf{f}^{\mathrm{spec}} \in \mathbb{R}^{k},
$$
which is similar to a discrete Fourier transform.
To project the function back to real space (inverse Fourier transform), we simply do
$$
\mathsf{f}' = \mathsf{Z} \mathsf{f}^{\mathrm{spec}}, \quad \mathsf{f}' \in \mathbb{R}^{N}.
$$

Resolution tuning is achieved by controlling the number of basis we use (i.e., tuning $k$) in spectral space.
As shown in \autoref{fig:fig1}, the electrostatic potential function resolution can be tuned by manipulating the number of
Laplace-Beltrami eigenfunctions.

However, what about the resolution of the shape itself? In \autoref{fig:fig1}, we see that the surface smoothness can also be tuned.
It is important to realize that manifold is an \textit{abstract} concept, which does not necessarily have a particular \textit{realization} 
in the Enclidean space. 
The surfaces that we visualize in \autoref{fig:fig1} are realizations of the underlying manifold in the Euclidean space,
where each surface vertex is associated with some Cartesian coordinates.
These coordinates are extrinsic properties of the manifold, thus can be treated in the same way as other surface functions (e.g., the electrostatic potential).
Therefore, in order to reconstruct the molecular surface with a lower spatial resolution, we can simply calculate the smoothed Cartesian coordinates:
$$
\mathsf{x}' = \mathsf{Z}\mathsf{x}^{\mathrm{spec}} = \mathsf{Z} \mathsf{Z}^{\top}\mathsf{B}\mathsf{x},
$$
and do the same for $\mathsf{y}$ and $\mathsf{z}$ to obtain the smoothed surface coordinates ($\mathsf{x}'$, $\mathsf{y}'$, $\mathsf{z}'$). 
In short, we realize resolution tuning with a series of (sparse) matrix multiplications
bringing the surface functions back-and-forth between the real space and the generalized Fourier space.

\setcounter{table}{0}
\setcounter{figure}{0}
\setcounter{equation}{0}

\section{Surface Preparation}\label{ap:surface_prep}

The raw input to the \model framework is simply the 3D atomic structures 
(e.g., \texttt{xyz} files for small molecules, or \texttt{PDB} files for proteins).
In other words, we only need to know where these atoms are and their atomic species.
For proteins with only heavy atoms (since hydrogen atoms are almost invisible under X-ray diffraction detectors),
we use \texttt{reduce}~\citep{word1999asparagine} 
(or alternatively \texttt{PDB2PQR}~\citep{dolinsky2004pdb2pqr}) software to add hydrogen atoms.

Next, we employ \texttt{MSMS}~\citep{ewing2010msms} to calculate the solvent-excluded surface of the molecule (with probe radius 1.5\,\AA, sampling density 3.0 for small molecules and 1.0 for proteins) as a triangle mesh.
We use \texttt{PyMesh}~\citep{zhou2019pymesh} to further refine the surface mesh in order to reduce the number of vertices and fix poorly meshed areas.
Degenerate vertices or disconnected surfaces would lead to numerical issues for solving the generalized eigenvalue problem in the next step, thus should be fixed beforehand.
We then compute the truncated Laplace-Beltrami eigenvectors, eigenvalues, and the mass matrix as described in \autoref{ap:harmonic_analysis}.

Initial geometric features can be directly calculated given the surface triangle mesh, 
where we use the \texttt{libigl}~\citep{jacobson2017libigl} package to calculate the mean and Gaussian curvatures,
and compute the Heat Kernel Signatures as described in~\cite{sun2009concise}.
These geometric features capture shape-related properties of the molecular surface, and are stored as a scalar-type array
$\mathsf{F}^{\mathrm{geom}} \in \mathbb{R}^{N \times p}$, where $p$ is the number of initial geometric features (a user defined variable).

Chemical features are projected from atoms to their neighboring surface vertices. 
We first obtain an initial descriptor vector $\bold{u}$ for each atom (e.g., atomic number, charge, etc.).
Then, for each surface vertex $\bold{x}_i$, we compute its $\nu$ nearest neighbor atoms centered at 
\{$\bold{a}_1^i$, $\dotsc$, $\bold{a}_{\nu}^i$\} with features \{$\bold{u}_1^i$, $\dotsc$, $\bold{u}_{\nu}^i$\}.
We apply a multilayer perceptron (MLP) to the vector [ $\bold{u}_{\nu}^i, 1/\|\bold{x}_i - \bold{a}_{\nu}^i\|$]
for each neighboring atom, then compute the average over the neighbors to obtain the chemical feature vector
$\mathsf{F}^{\mathrm{chem}} \in \mathbb{R}^{N \times q}$, 
where $q$ is a user defined variable indicating the dimension of initial chemical features.
In short, the initial chemical features of each surface vertex are determined by
its neighboring atomic species and their distance, which is learned by a MLP.

We use another MLP to combine the per-vertex initial features 
$$
\mathsf{F}^{\mathrm{inp}} \leftarrow \mathrm{MLP} (\mathrm{concat}(\mathsf{F}^{\mathrm{geom}}, \mathsf{F}^{\mathrm{chem}})),\quad \mathsf{F}^{\mathrm{inp}} \in \mathbb{R}^{N \times n}.
$$
These $n$ features reflect the local geometric and chemical environment of each surface vertex, 
which will be used as the input to the harmonic message passing module.

The output of the surface preparation module includes (1) the truncated Laplace-Beltrami eigenvectors $\mathsf{Z} \in \mathbb{R}^{N \times k}$, the corresponding eigenvalues $\{\lambda_i\}_{i=0}^{k-1}$, and the sparse mass matrix $\mathsf{B} \in \mathbb{R}^{N \times N}$,
(2) per-vertex scalar features $\mathsf{F}^{\mathrm{inp}} \in \mathbb{R}^{N \times n}$.

\setcounter{table}{0}
\setcounter{figure}{0}
\setcounter{equation}{0}

\section{The Binding Site Prediction Module}\label{sec:details_of_BSP}

First, we define the binding site as the protein surface region which is within 3\,$\si{\angstrom}$ to its counterpart surface, and obtain the set of all corresponding surface points $P$ as the nearest neighbor vertices between the ground truth protein interfaces.

The binding site prediction module stacks two feature propagation blocks, each consists of three \model layers (introduced in \autoref{sec:HMP}) and a cross attention layer. Given the propagated receptor features $\mathsf{F}$ and ligand features $\mathsf{G}$, the cross attention layer enables communication between proteins:
\begin{equation*}
\begin{aligned}
    \mathsf{F'} &= 
    \mathrm{softmax}\left( \frac{(\mathsf{F} \mathsf{W}_\mathrm{Q}) (\mathsf{G} \mathsf{W}_\mathrm{K})^\top}{\sqrt{d_k}}\right) (\mathsf{G} \mathsf{W}_\mathrm{V}),\\
    \mathsf{G'} &= \mathrm{softmax}\left( \frac{(\mathsf{G} \mathsf{W}_\mathrm{Q}) (\mathsf{F} \mathsf{W}_\mathrm{K})^\top}{\sqrt{d_k}}\right) (\mathsf{F} \mathsf{W}_\mathrm{V}),
\end{aligned}
\end{equation*}
where $\mathsf{F}\in \mathbb{R}^{N_R \times d_k}$, $\mathsf{G} \in \mathbb{R}^{N_L \times d_k}$ denotes the propagated features on the receptor/ligand protein surface, $d_k$ denotes the dimension of features,
and $\mathsf{W}_\mathrm{Q}$, $\mathsf{W}_\mathrm{K}$ and $\mathsf{W}_\mathrm{V}$ are the parameter matrices for the query, key, and value in attention computation, respectively.

The loss function consists of two components. The first is a binary cross entropy loss, which encourages the model to correctly predict the binding site:
\begin{equation*}
    \mathcal{L}_{bce}(i) = -[ y_i \log x_i + (1- y_i) \log (1-x_i)],
\end{equation*}
where $y_i$ and $x_i$ are the label and predicted probability of whether vertex $i$ belongs to the binding site.

The second term is a PointInfoNCE loss \citep{xie2020pointcontrast}, a contrastive matching loss that minimizes the distance between the features of corresponding surface points and  maximizes  the distance between non-corresponding point features:
\begin{equation*}
    \mathcal{L}_{match} = - \sum_{(i, j) \in P} \log \frac{\exp(\mathbf{f}_i \cdot \mathbf{g}_j / \,\tau)}{\sum_{(\cdot, k)\in P}\exp(\mathbf{f}_i \cdot \mathbf{g}_k / \,\tau)} \,,
\end{equation*}
where $P$ is the set of all corresponding surface points and $\tau$ is the temperature factor (a hyperparameter). 
Here $\mathbf{f}_i$ and $\mathbf{g}_j$ are the neural network-learned feature vectors at point $i$ and $j$, 
which belong the the receptor and ligand surface, respectively.

The total loss is the weighted sum of the two loss terms:
\begin{equation*}
    \mathcal{L} = \mathcal{L}_{bce} + \lambda \mathcal{L}_{match} \,,
\end{equation*}
where empirically we set $\lambda$ to 0.1 in our docking experiments.

\setcounter{table}{0}
\setcounter{figure}{0}
\setcounter{equation}{0}

\section{Details on Functional Maps}\label{ap:fmaps}
In this section, we present the details of functional maps used in~\autoref{sec:method_docking}. 

Let us be given two manifolds $\mathcal{M}$ and $\mathcal{N}$. 
The aim of functional maps is to find a bijective mapping $T:\mathcal{M} \to \mathcal{N}$ to align these two manifolds.
Unlike traditional methods that try to recover the point-to-point correspondence directly, functional maps lift the mapping $T$ to a correspondence between the functional spaces on the two manifolds.
Formally, let $L^2(\cdot)$ be the functional space of square integrable functions on a manifold, we infer the functional correspondence $T_F: L^2(\mathcal{M}) \to L^2(\mathcal{N})$ induced by the mapping $T$, and is defined by $T_F f = f \circ T^{-1}$ for any $f\in L^2(\mathcal{M})$. 

To compute the functional correspondence, we need to utilize the Laplace-Beltrami basis on the manifolds. 
Actually, such functional correspondence has a concise expression in spectral domain: 
given the respective truncated LB eigenfunctions $\{\phi_j^\mathcal{M}\}_{j\geq0}^{k_1}$ on $\mathcal{M}$ and $\{\phi_i^\mathcal{N}\}_{i\geq0}^{k_2}$ on $\mathcal{N}$,
the functional correspondence $T_F$ can be (approximately) represented as a change of basis matrix:
\[\mathsf{C} = (c_{ij})_{k_2\times k_1} = (\langle \phi_i^\mathcal{N}, T_F \phi_j^\mathcal{M} \rangle_{\mathcal{N}})_{k_2\times k_1}.
\]
Now given a set of $q$ corresponding functions $\{f_1, \dots, f_q\} \subset L^2(\mathcal{M})$ and $\{g_1, \dots, g_q\} \subset L^2(\mathcal{N})$, we denote their spectral representations by coefficients $\mathsf{A} = (a_{ij})_{k_1\times q}$, where $a_{ij} = \langle \phi_i^\mathcal{M}, f_j \rangle_{\mathcal{M}}$, and $\mathsf{B} = (b_{ij})_{k_2 \times q}$, where $ b_{ij} = \langle \phi_i^\mathcal{N}, g_j \rangle_{\mathcal{N}}$.
The matrix $\mathsf{C}$ can be obtained by solving the following quadratic minimization problem:
\begin{equation}\label{eqn:fmap_opt}
    \mathsf{C} = \mathrm{argmin}_{\mathsf{C} \in \mathbb{R}^{k_2\times k_1}} \|\mathsf{C} \cdot \mathsf{A} - \mathsf{B}\|_\mathrm{F}^2 
    + \alpha \|\mathsf{C} \cdot \mathsf{\delta}_\mathcal{M} - \mathsf{\delta}_\mathcal{N} \cdot \mathsf{C}\|_\mathrm{F}^2 
    + \beta \sum_{i=1}^q \|\mathsf{C} \cdot \mathsf{\Lambda}_{\mathcal{M}}^i - \mathsf{\Lambda}_{\mathcal{N}}^i \cdot \mathsf{C}\|_\mathrm{F}^2 \,,
\end{equation}
where $\|\cdot\|_\mathrm{F}$ denotes the Frobenius norm, and the first term on the right hand side is the change of basis constraint. Two more regularization terms are introduced into the formula with tunable weights $\alpha, \beta > 0$. $\|\mathsf{C} \cdot \mathsf{\delta}_\mathcal{M} - \mathsf{\delta}_\mathcal{N} \cdot \mathsf{C}\|_\mathrm{F}$ enforces the isometry of the two manifolds where the matrices $\delta_\mathcal{M}$ and $\delta_\mathcal{N}$ are the spectral representation of the LB operators. The matrices $\mathsf{\Lambda}_\mathcal{M}^i$ and $\mathsf{\Lambda}_\mathcal{N}^i$ are called the orientation operator~\citep{Re:2018continuous} and are defined by the functions $f_i$ and $g_i$, respectively. The commutator with the orientation operator incorporates extrinsic properties into the formulation and enforces the orientation of functional maps (i.e., which side of the 2D surface is pointing ``outward'').

Once the functional mapping (i.e., the $\mathsf{C}$ matrix) is recovered, point-to-point correspondence $T$ between manifold $\mathcal{M}$ and $\mathcal{N}$ can be obtained by mapping indicator functions of vertices on $\mathcal{M}$ to the corresponding functions on $\mathcal{N}$ because $T_F \delta_m = \delta_{T(m)}$ for any vertex $m\in\mathcal{M}$~\citep{Ov:2012functional}.

\setcounter{table}{0}
\setcounter{figure}{0}
\setcounter{equation}{0}

\section{Details on QM9 Property Regression}\label{ap:qm9}

\textbf{Dataset}\quad 
For the QM9 molecular property regression task, we align our input data with EGNN~\citep{Sa:2021n} (with a total of 130,831 instances).
181 molecules failed during molecular surface extraction (\texttt{MSMS} failed to generate reasonable surface mesh for some molecules).
We follow the same data split as EGNN, leading to 130,650 instances (99,862 for training, 17,719 for validation, and 13,069 for test).
Only the atomic number and atomic positions are used as the initial molecular information, no extra handcrafted features are involved.
We compute the molecular surface as described in \autoref{ap:surface_prep}, 
with an average of 439 vertices and 47 eigenfunctions per molecular surface.  

\textbf{Model Architecture and Performance}\quad
Different from other graph-based models, we do not explicitly encode inter-atomic distance or chemical bonding information.
Instead, we feed the molecular surface manifold as the model input, 
which contains local chemical information (\autoref{ap:surface_prep}).
Surface geometric features (i.e., curvature and the Heat Kernel Signatures) are not included, 
since we find these features have no contribution to predicting molecular properties.
We stack 6 layers of \model, followed by a global average pooling layer to aggregate information from all surface points
to make a final property prediction. 
Batch normalization is applied to all MLPs.
The mean absolute prediction error of all 12 properties are shown in \autoref{tab:qm9_all}, 
results averaged over three independent runs with different random seeds.

\begin{table}[H]
\footnotesize
\setlength{\tabcolsep}{3pt}
\caption{Model performance on the QM9 dataset, reporting the Mean Absolute Error (MAE).}
\label{tab:qm9_all}
\begin{center}
\begin{tabular}{lcccccccccccc}
\toprule
Task & $\alpha$ & $\Delta$ $\varepsilon$ & $\varepsilon_{\mathrm{HOMO}}$ & $\varepsilon_{\mathrm{LUMO}}$ & $\mu$ & $C_{\nu}$ &
$G$ & $H$ & $r^2$ & $U$ & $U_0$ & ZPVE \\
\midrule
Units & bohr$^3$ & meV & meV & meV & D & cal/(mol\,K) & meV & meV & bohr$^2$ & meV & meV & meV \\
\midrule
NMP & .092 & 69 & 43 & 38 & .030 & .040 & 19 & 17 & .180 & 20 & 20 & 1.50\\
SchNet & .235 & 63 & 41 & 34 & .033 & .033 & 14 & 14 & .073 & 19 & 14 & 1.70\\
Cormorant & .085 & 61 & 34 & 38 & .038 & .026 & 20 & 21 & .961 & 21 & 22 & 2.02\\
SE(3)-Tr. & .142 & 53 & 35 & 33 & .051 & .054 & -- & -- & -- & -- & -- & --\\
EGNN & .071 & 48 & 29 & 25 & .029 & .031 & 12 & 12 & .106 & 12 & 12 & 1.55\\
SEGNN & .060 & 42 & 24 & 21 & .023 & .031 & 15 & 16 & .660 & 13 & 15 & 1.62\\
\midrule
\textbf{\model} & .102 & 59 & 37 & 31 & .037 & .040 & 40 & 41 & .653 & 44 & 42 & 2.97\\
\bottomrule
\end{tabular}
\end{center}
\end{table}

\textbf{Resource Consumption}\quad
Computing the molecular surface and its Laplace-Beltrami eigenfunctions is computationally intensive.
We performed data preprocessing in parallel on 64 CPUs, where
molecular surface computation took 198 seconds, surface mesh refinement took 59 minutes,
and solving eigenfunctions took 4.5 hours.
With an average of 439 mesh vertices and 47 eigenfunctions, the entire QM9 dataset consumes 12.1\,GB disk space 
(approximately 100\,kB per molecule).
The model for production contains 356,609 learnable parameters, most of which are linear transformation coefficients in MLPs.
However, manifold harmonic message passing involves matrix multiplications bringing features back-and-forth between the real space
and the generalized Fourier space, which is performed in a serial manner (instead of batch processing).
This is because the number of surface vertices and Laplace-Beltrami eigenfunctions are different across different molecules.
Therefore, we only perform batch operations on the MLPs, but not on message passing layers.
We trained our model on NVIDIA A100 GPUs with 80\,GB memory with a batch size of 32, which on average takes 
240 seconds to train a single epoch (99,862 molecules), and 16 seconds for inference on the test set (13,069 molecules).
For each molecule we keep its $N \times n$ feature matrix $\mathsf{F}$ ($N$ vertices and $n$ features),
and the $N \times k$ eigenvectors ($\mathsf{Z}$ matrix) and its inverse matrix ($\mathsf{Z}^{\top}\mathsf{B}$ matrix, see \autoref{ap:harmonic_analysis}), 
which does not consume much GPU memory.

\setcounter{table}{0}
\setcounter{figure}{0}
\setcounter{equation}{0}

\section{Details on Ligand-binding Pocket Classification}\label{ap:masif-ligand}

\textbf{Dataset}\quad The dataset is obtained from \cite{Ga:2020deciphering} with 1,438 non-redundant protein structures that each corresponds to a list of bound ligands with their atom coordinates.
We first generate protein surface meshes as described in \autoref{ap:surface_prep}. To extract ligand binding pockets, we identify pocket vertices on the surface mesh that are within 4 \si{\angstrom} to any atom of the ligand and extract the largest connected component of pocket vertices as the binding pocket. Binding pockets that contain $< 100$ vertices are removed and the LB eigenfunctions for the remaining pocket surfaces are calculated. 3 protein complexes failed in protein surfaces generation and 63 protein complexes failed in binding pocket extraction due to disconnected surface or too few vertices after surface refinement.
In total, 2,254 binding pockets are extracted that bind to one of seven ligands: ADP (634), HEM (396), FAD (338), COA (263), NAD (232), SAM (222), and NAP (169). On average, 1,199 vertices and 190 eigenfunctions per pocket surface are obtained. 
The same training, validation, and test split as \cite{Ga:2020deciphering} are used and we obtained 1,634 training pockets (in 986 protein complexes), 202 validation pockets (in 112 protein complexes), and 418 test pockets (in 274 protein complexes). Similar to \cite{Ga:2020deciphering}, we provide chemical information (hydrophobicity score, partial charge) as well as local geometric information (mean/Gaussian curvature and the Heat Kernel Signatures) as input features.

\begin{figure}[h]
\centering
\includegraphics[width=0.5\textwidth]{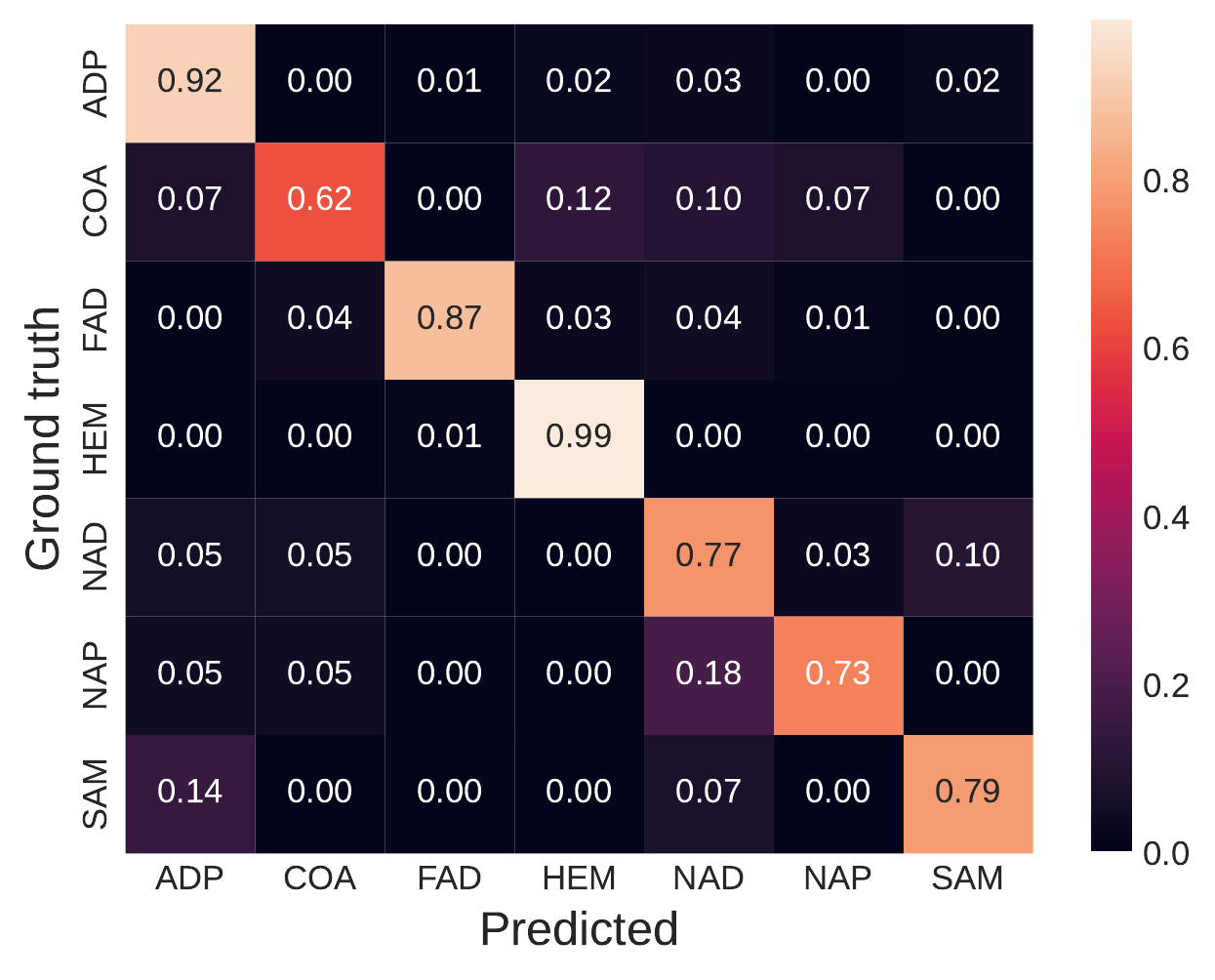}
  \caption{Confusion matrix of ligand specificity predicted by \model on the test set.
  }\label{fig:figa2}
\end{figure}

\textbf{Model Architecture and Performance}\quad
The \model-based classification model contains 6 layers of \model propagation layers followed by a global average pooling to aggregate information from all surface points of a pocket. A simple 2-layer MLP is used to classify pockets into seven ligand-binding classes. \model is trained to minimize the cross-entropy loss for 400 epochs (approx. 20,000 iterations with a batch size of 32) and the one with the best balanced accuracy score on the validation set is selected. To compare with MaSIF-ligand, we calculated the balanced accuracy for multi-class classification (\autoref{fig:fig6}). 
Per class performance of our full model (geometric + chemical features) is shown as a confusion matrix (\autoref{fig:figa2}).

\textbf{Resource Consumption}\quad
Similar to QM9, we performed data preprocessing in parallel on 64 CPUs. For all 2,254 protein pocket cases: the protein surface computation took 306 seconds, surface pocket extraction and refinement took 330 seconds, and solving the eigenfunctions for the extracted pockets took 26 minutes. The processed dataset (N=2,254) takes 2.4\,GB disk space (about 1.1 MB per protein pocket). The full model contains 407,047 trainable parameters. 
We trained our model on NVIDIA A100 GPUs with 80 GB memory with a batch size of 32, which on average takes
33 seconds to train a single epoch (1,634 protein pockets), and 9 seconds for inference on the test set
(418 protein pockets). The model structure is almost identical to QM9 that shares the same batch processing design and has similar GPU memory consumption.

\setcounter{table}{0}
\setcounter{figure}{0}
\setcounter{equation}{0}

\section{Details on Rigid Protein Docking} \label{ap:docking}

\subsection{Dataset}

The original DIPS dataset~\citep{townshend_dips_2019} contains interacting protein chains extracted from experimentally determined complex structures in the RCSB PDB database (\url{https://www.rcsb.org/}). While DIPS is designed to represent the interactions in protein complexes including homomeric proteins (i.e., protein complexes composed of identical proteins), we are particularly interested in interactions between different proteins in biological processes (e.g., antibody-antigen, enzyme-inhibitor, enzyme-substrates, etc.). To better resemble such interactions, we apply a modified pipeline based on DIPS, referred as DIPS-Het.

Specifically, we include the PDB entries in DIPS as well as newly deposited entries from the RCSB PDB database that (1) are determined using diffraction-based methods or electron microscopy, (2) have $<$ 3.5 \si{\angstrom} resolution, (3) are not hybrid protein complexes (e.g., protein-DNA complexes are excluded), and (4) have $<$ 5 protein chains. Different from the original DIPS pipeline, we a) further remove the PDB entries with only homomeric interfaces (i.e., interactions between identical proteins), and b) adopt a one-vs-rest strategy that uses each one of the protein chains in the complex as the ``ligand protein'' and the rest protein chains as the ``receptor protein'' to form ligand-receptor pairs.

Similar to \cite{Ga:2021independent}, we remove the cases with any protein chain sharing the same protein sequence clusters (at 30\% sequence similarity) with DB5.5. Finally, we cluster proteins based on protein chain sequence similarity and separate these clusters into the training and validation set. This sequence-based split helps selecting a model with better generalizability. After removing proteins failed in surface mesh generation, DIPS-Het contains 11,373 training cases and 508 validation cases.

\subsection{Evaluation Metrics}
We evaluate the rigid protein docking results using four metrics: Complex RMSD, Interface RMSD, DockQ, and Success rate. Complex RMSD and Interface RMSD are used in \cite{Ga:2021independent}: let $\mathbf{Z}^* \in \mathbb{R}^{3 \times (n + m)}$ and $\mathbf{Z} \in \mathbb{R}^{3\times(n + m)}$ be the $\alpha$-carbon coordinates of the ground truth and predicted protein complexes, respectively, 
were $m$ and $n$ are the number of $\alpha$-carbons in the receptor and ligand protein.
After superimposing the complex structures using the Kabsch algorithm, Complex RMSD is calculated as $\sqrt{\frac{1}{n + m}|| \mathbf{Z}^* - \mathbf{Z}||^2_F}$. 
Similarly, Interface RMSD is calculated using the same procedure but with the $\alpha$-carbon coordinates of interface residues ($<$ 8 \si{\angstrom} to the other protein's residues). Smaller RMSD value means the predicted structure is closer the ground truth structure.

In addition, we evaluate the overall quality of docking using DockQ~\citep{basu_dockq_2016} and the success rate of achieving ``Acceptable'' or higher according to ~\cite{mendez_assessment_2003,mendez_assessment_2005}. Both metrics are based on three standardized criteria used by Critical Assessment of PRedicted Interactions (CAPRI): $L_\text{rms}$ is the ligand (the smaller protein) RMSD calculated based on backbone atoms, after superimposing receptor's backbone atoms; $I_\text{rms}$ is the backbone RMSD of interface residues, after superimposing the interface residues (residues with any atom is $<$ 10 \si{\angstrom} to atoms in the other protein); and $f_\text{nat}$ is the recall in recovering residue-residue contacts between the proteins, where two residues are ``in contact'' if any pair of atoms from two residues has distance $<$ 5 \si{\angstrom}. DockQ is a continuous score between 0 and 1 (the higher the better), derived from $L_\text{rms}$, $I_\text{rms}$, and $f_\text{nat}$. A docking result is considered as a ``success'' if it is ranked ``Acceptable'' or higher according to CAPRI's criteria, that is
\begin{align*}
f_\text{nat} \ge 0.1 \land (L_\text{rms} &\le 10.0 \lor I_\text{rms} \le 4.0) \\
& \text{OR} \\
f_\text{nat} &\ge 0.3
\end{align*}
Both DockQ score and CAPRI ranking is calculated using the DockQ package (\url{https://github.com/bjornwallner/DockQ/}).

\subsection{Resource Consumption}

\textbf{Dataset}\quad
We report the rigid protein docking resource consumption using the Docking Benchmark 5.5 (DB5.5) dataset~\citep{guest_expanded_db55_2021},
which contains 253 pairs of representative protein complex structures. 
This is the gold-standard test set to evaluate protein docking model performance.

We preprocessed the DB5.5 dataset in parallel on 64 CPUs. 
Adding hydrogen atoms to the PDB data took 34 seconds (using \texttt{reduce}), 
computing the solvent-excluded molecular surface with \texttt{MSMS} took 6 seconds,
triangle mesh refinement by \texttt{PyMesh} took 3.4 minutes, 
and computing the Laplace-Beltrami eigenfunctions took 18 minutes with \texttt{scipy} (\texttt{eigsh})\footnote{Solving the generalized eigenvalue problem should be done with less parallel processes to allocate more CPU resources to each solver for better efficiency.}.
The average number of vertices per protein (i.e., ligand or receptor) is 3,380, with 215 calculated Laplace-Beltrami eigenfunctions (maximum eigenvalue capped to 0.3, which is determined empirically). 
The total storage space for the processed dataset is 2\,GB (about 8\,MB per protein, much larger than QM9 molecules). For our training set with 11,781 protein complexes, the storage space is 101\,GB.

\textbf{Inference}\quad
The \model inference time in predicting DB5.5 proteins complexes is shown in \autoref{tab:inference_time}, 
in comparison to \textsc{Equidock}, \textsc{Hdock}, and \textsc{Attract}.
For \model, data preprocessing took 80\% of inference time, functional maps (docking pose prediction) took 19\%,
while binding site prediction took only 1\% of time.

\begin{table}[H]
\footnotesize
\setlength{\tabcolsep}{6pt}
\caption{Rigid protein docking inference time on DB5.5 dataset (per protein complex docking time averaged over 253 cases).}
\label{tab:inference_time}
\begin{center}
\begin{tabular}{cc}
\toprule
Method & Time \\
\midrule
HDock & 884.9 sec \\
ATTRACT & 882.7 sec \\
EquiDock & 3.9 sec \\
\textbf{\model} (Top 1) & 5.8 sec \\
\textbf{\model} (Top 3) & 6.6 sec \\
\bottomrule
\end{tabular}
\end{center}
\label{tab:docking_time}
\end{table}

\subsection{Distributions of learned propagation parameter}

\begin{figure}[H]
  \centering
  \includegraphics[width=0.8\textwidth]{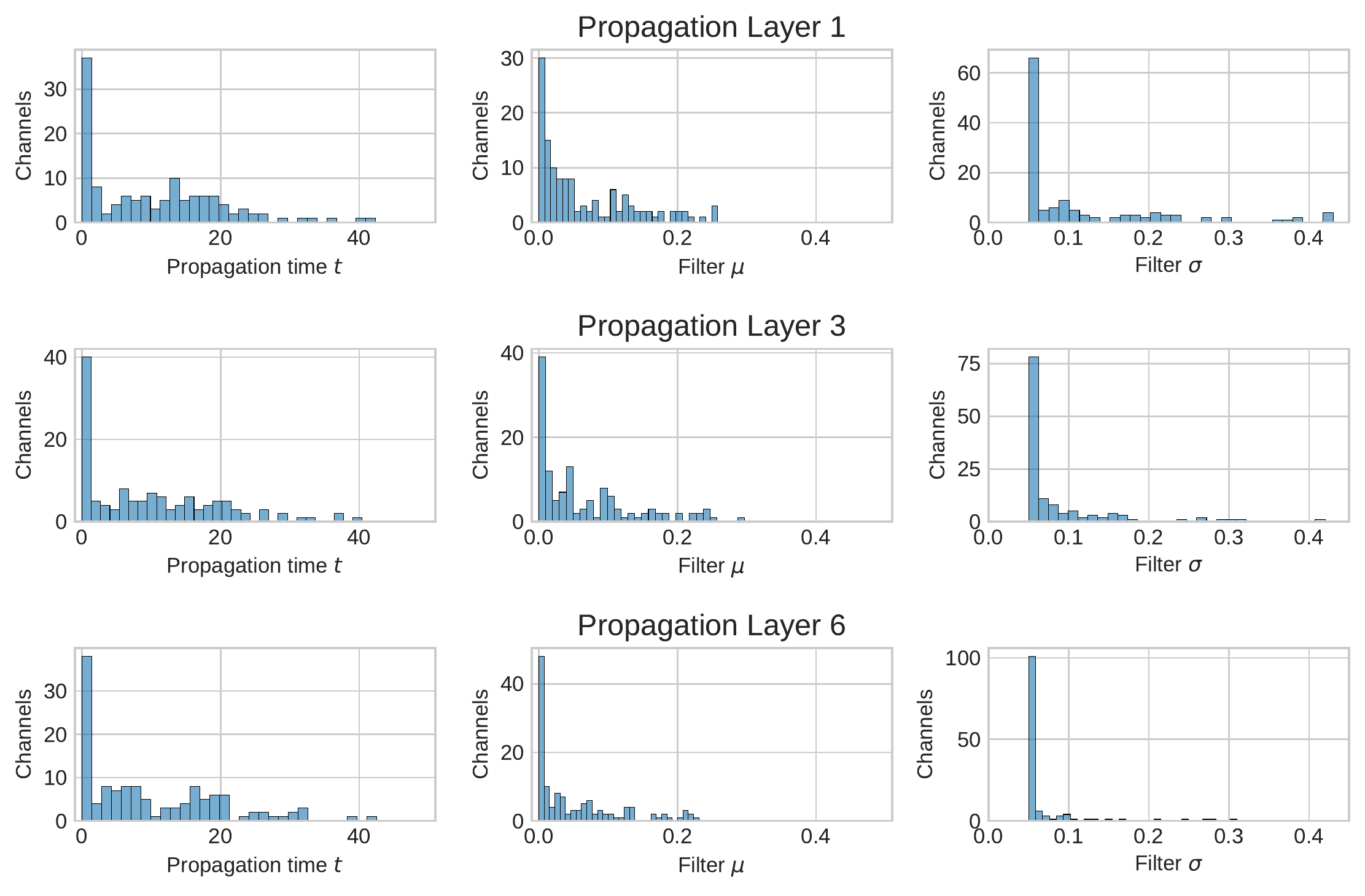}
  \caption{Distribution of learned propagation time $t$ and frequency filter ($\mu$ and $\sigma$) at selected propagation layers.}
  \label{fig:figa3}
\end{figure}

We visualize the distribution of propagation parameters: propagation time $t$, frequency filter mean $\mu$ and variance $\sigma$, learned from data. The histogram of selected layers are shown in \autoref{fig:figa3}.

\subsection{Analysis of Different Message Passing Mechanisms}

We study the effectiveness of proposed Harmonic Message Passing mechanism by comparing it with spatial graph message passing, a ``Graph Message Passing'' model where Harmonic Massage Passing is replaced by Graph Attention Network \citep{velickovic_graph_2018}, and with a ``No Message Passing'' model where only a multilayer perceptron (MLP) is used to encode per-vertex features.
Experimental results show that the Harmonic Message Passing model outperforms Graph Message Passing and No Message Passing models, suggesting its higher efficiency in propagating information on protein surface meshes.

\begin{table}[H]
\setlength{\tabcolsep}{4pt}
\caption{Rigid Protein Docking Performance with Different Message Passing Models}
\label{tab:mp-mech}
\begin{center}
\begin{tabular}{lccccccc}
\toprule
\multicolumn{1}{l}{}             & \multicolumn{2}{c}{Complex RMSD} & \multicolumn{2}{c}{Interface RMSD}  & \multicolumn{2}{c}{{ DockQ}}  \\ \cmidrule(lr){2-3} \cmidrule(lr){4-5} \cmidrule(lr){6-7}
\multicolumn{1}{l}{Model}        & \multicolumn{1}{c}{{ Median}} & \multicolumn{1}{c}{{ Mean}}  & \multicolumn{1}{c}{{ Median}} & \multicolumn{1}{c}{{ Mean}}  & \multicolumn{1}{c}{{ Median}} & \multicolumn{1}{c}{{ Mean}}   \\
\midrule
\textbf{Harmonic Message Passing} & \bf 12.01 &\bf 12.09 &\bf 10.90 &\bf 11.43 &\bf 0.06 &\bf 0.22 \\
Graph Message Passing& 15.41&15.67&13.38&14.37&0.03&0.08 \\
No Message Passing & 16.44&17.23&14.11&16.11&0.02&0.04 \\
\bottomrule
\end{tabular}
\end{center}
\end{table}

\subsection{Ablation Studies on Input Features}

To assess the importance of input chemical and geometric features for rigid protein docking, we compare the performance of \model (``Full model") to models trained with only chemical or geometric features: the ``Chem only'' model only includes chemical features (i.e., atom types, residue types, residue's hydrophobicity, and whether the atom is an $\alpha$-carbon); the ``Geom only'' model only contains features related to manifold geometry, including Gaussian curvature, mean curvature, and Heat Kernel Signatures. \autoref{tab:feat-ablation} shows that the model performance drops upon removing either chemical or geometric features. And the chemical features is more critical for \model in the rigid protein docking task.

\begin{table}[H]
\setlength{\tabcolsep}{2.5pt}
\caption{Ablation studies on input features}
\label{tab:feat-ablation}
\begin{center}
\begin{tabular}{lccccccc}
\toprule
{} & \multicolumn{2}{l}{Complex RMSD} & \multicolumn{2}{l}{Interface RMSD} & \multicolumn{2}{c}{DockQ} \\ \cmidrule(lr){2-3} \cmidrule(lr){4-5} \cmidrule(lr){6-7}
Model      &                 Median &   Mean &         Median &   Mean     &   Median &  Mean              \\ \midrule
\textbf{Full model} &        \bf 12.01 & \bf  12.09   & \bf 10.90 & \bf 11.43     & \bf 0.06 & \bf 0.22    \\
Chem only  &                  13.96 &  13.95 &          12.73 &  13.33     &     0.04 &  0.17              \\
Geom only  &                  17.87 &  19.15 &          16.58 &  18.46     &     0.01 &  0.03              \\
\bottomrule
\end{tabular}
\end{center}
\end{table}

\subsection{Cross experiments with \textsc{EquiDock}}

Compared to \textsc{EquiDock}, \model differs both in the model and the dataset used for training. 
To understand which aspect contributes more to the performance of \model, we conduct cross experiments to train \model on the original DIPS (used in \textsc{EquiDock}) and \textsc{EquiDock} trained on DIPS-Het (proposed by us, used in \model). 
The same training and validation sets are used, and all models are evaluated on DB5.5. As summarized in \autoref{tab:cross-exp}, model structure and training dataset are both important: \model achieved higher performance than \textsc{EquiDock} when both models are trained on DIPS dataset; the performance of \model further improves when it is trained on DIPS-Het dataset. 

\begin{table}[H]
\setlength{\tabcolsep}{2.5pt}
\caption{Compare \textsc{EquiDock} and \model trained on DIPS or DIPS-Het datasets}
\label{tab:cross-exp}
\begin{center}

\begin{tabular}{lcccccccc}
\toprule
                                    &                 & \multicolumn{2}{c}{Complex RMSD} & \multicolumn{2}{c}{Interface RMSD} & \multicolumn{2}{c}{DockQ} \\ \cmidrule(lr){3-4} \cmidrule(lr){5-6} \cmidrule(lr){7-8}
Model                               & Training data   &                 Median &   Mean &           Median &   Mean &  Median &  Mean \\ \midrule
\multirow{2}{*}{\model}             & DIPS            &                  15.68 &  16.04 &            14.81 &  15.38 &    0.02 &  0.07 \\
                                    & DIPS-Het        &                  12.01 &  12.09 &            10.90 &  11.43 &    0.06 &  0.22 \\ \midrule
\multirow{2}{*}{\textsc{EquiDock}}  & DIPS            &                  17.27 &  18.09 &            15.04 &  16.11 &    0.02 &  0.03 \\
                                    & DIPS-Het        &                  17.22 &  17.65 &            14.27 &  14.76 &    0.02 &  0.04 \\
\bottomrule
\end{tabular}

\end{center}
\end{table}

\subsection{DB5.5 test cases from \textsc{EquiDock}}

For a fair comparison, we further compare \model with \textsc{EquiDock} on the 25 cases in DB5.5 selected as a test set in \cite{Ga:2021independent}. 
We report the \textsc{EquiDock} model fine-tuned on DB5.5, as originally reported in \cite{Ga:2021independent}. 
As shown in \autoref{tab:test-set-bench}, \model still outperforms \textsc{EquiDock}, despite not being fine-tuned on DB5.5.

\begin{table}[H]
\setlength{\tabcolsep}{2.5pt}
\caption{Rigid docking performance on DB5.5 test set}
\label{tab:test-set-bench}
\begin{center}

\begin{tabular}{lcccccccccc}
\toprule
{} & \multicolumn{2}{c}{Complex RMSD} & \multicolumn{2}{c}{Interface RMSD} & \multicolumn{2}{c}{DockQ} & Success rate\\  \cmidrule(lr){2-3} \cmidrule(lr){4-5} \cmidrule(lr){6-7}
Model             &       Median &   Mean &         Median &   Mean &   Median &  Mean &                        ($\geq$ Acceptable) \\ \midrule
\model (Top 1)    &        13.10 &  12.42 &          11.05 &  11.68 &     0.07 &  0.15 &                         0.20 \\
\model (Top 3)    &        11.35 &  11.18 &          10.02 &  10.03 &     0.08 &  0.16 &                         0.20 \\
\textsc{EquiDock} &        14.13 &  14.72 &          11.97 &  13.23 &     0.04 &  0.05 &                         0.00 \\
\bottomrule
\end{tabular}

\end{center}
\end{table}

\subsection{Model performance at different resolutions}

We perform the rigid docking experiment using \model at different resolutions. Specifically, we cap the largest eigenvalue
of the ligand and receptor protein surface manifolds used in the model. Since eigenfunctions with larger eigenvalues exhibit higher 
spatial resolutions, restricting the eigenvalues effectively constrains the model resolution. The results are shown in \autoref{tab:resolution_tuning}. In general, we observe worse model performance with fewer eigenfunctions for harmonic message passing and functional maps. Interestingly, the metrics of the binding site prediction module (i.e., AUC and AP) are less sensitive to resolution than the rigid docking module (powered by functional maps). This means the model could still infer where the binding site is with lower resolution, yet the quality of learned functional correspondence decreases, leading to worse docking power.

\begin{table}[H]
\footnotesize
\setlength{\tabcolsep}{6pt}
\caption{\model rigid protein docking performance at different spatial resolutions (quantified by the largest eigenvalue of surface manifold). AUC (area under the receiver operating characteristic curve) and AP (average precision) are metrics of the binding site prediction module.}
\label{tab:resolution_tuning}
\begin{center}
\begin{tabular}{ccccc}
\toprule
max eigenvalue & AUC & AP & CRMSD (Top 1) & CRMSD (Top 3) \\
\midrule
0.30 & 0.84 & 0.54 & 12.11 & 8.36 \\
0.25 & 0.84 & 0.53 & 13.14 & 8.9 \\
0.20 & 0.84 & 0.54 & 11.91 & 9.53 \\
0.15 & 0.83 & 0.53 & 12.19 & 9.91 \\
0.10 & 0.82 & 0.52 & 12.90 & 10.02 \\
0.05 & 0.81 & 0.51 & 13.84 & 11.32 \\
\bottomrule
\end{tabular}
\end{center}
\end{table}

\subsection{Training details}\quad
%

 \begin{table*}[htbp]
\centering
\caption{Hyperparameter choices of \model and the training phase settings}
\setlength{\tabcolsep}{16mm
}
\begin{tabular}{ll}
\toprule[1.2pt]

\textbf{Hyperparameters} & \textbf{Values}\\ 
 \midrule[0.5pt]
 \multicolumn{2}{c}{\textbf{Binding Site Prediction Module}} \\
 \midrule[0.5pt]
Number of Feature Propagation Blocks &  2  \\
Number of \model Layers &  3  \\
Dimension of Propagated Features $d_\textit{k}$& 128  \\ 
Number of Attention Heads & 4  \\
Dropout Rate & 0.1 \\
NCELoss Temperature & 10 \\
NCELoss Number of Point Samples & 50 \\
NCELoss Weight $\lambda$ & 0.1 \\
\midrule[0.5pt]
\multicolumn{2}{c}{\textbf{Training}} \\
\midrule[0.5pt]
Batch Size  &  4 \\
Epoch & 80  \\
Learning Rate   & 5$\times 10^{-4}$  \\
Optimizer  &  Adam  \\
Learning Rate Scheduler & Cosine Annealing  \\
\bottomrule[1.2pt]
\end{tabular}
\label{tab:hyperparameters}
\end{table*}

\vspace{0pt}
\begin{figure}[H]
  \centering
  \includegraphics[width=0.7\textwidth]{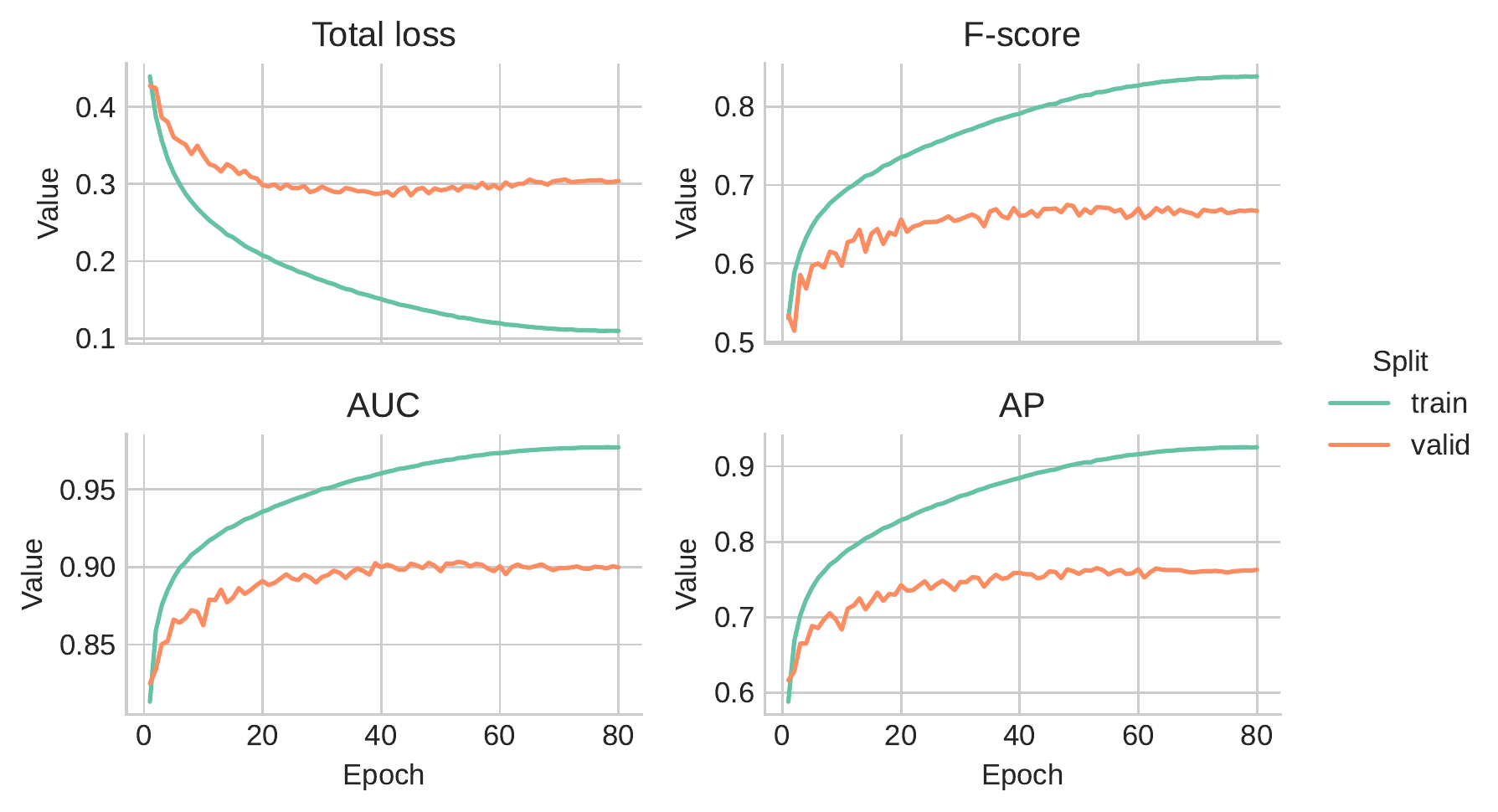}
  \caption{\model learning curve for the rigid protein docking task. Model with the best validation average precision (AP) score in binding site prediction is selected for testing on DB5.5.}
  \label{fig:figa4}
\end{figure}

\end{document}

%% file: math_commands.tex
\usepackage{amsmath,amsfonts,bm}

\def\eqref#1{equation~\ref{#1}}

\def\1{\bm{1}}

\DeclareMathAlphabet{\mathsfit}{\encodingdefault}{\sfdefault}{m}{sl}
\SetMathAlphabet{\mathsfit}{bold}{\encodingdefault}{\sfdefault}{bx}{n}

\DeclareMathOperator*{\argmin}{arg\,min}

%% file: iclr2023_conference.bbl
\begin{thebibliography}{72}
\providecommand{\natexlab}[1]{#1}
\providecommand{\url}[1]{\texttt{#1}}
\expandafter\ifx\csname urlstyle\endcsname\relax
  \providecommand{\doi}[1]{doi: #1}\else
  \providecommand{\doi}{doi: \begingroup \urlstyle{rm}\Url}\fi

\bibitem[Anderson et~al.(2019)Anderson, Hy, and Kondor]{An:2019cormorant}
Brandon Anderson, Truong~Son Hy, and Risi Kondor.
\newblock Cormorant: {C}ovariant molecular neural networks.
\newblock \emph{Advances in neural information processing systems}, 32, 2019.

\bibitem[Attaiki et~al.(2021)Attaiki, Pai, and Ovsjanikov]{At:2021dpfm}
Souhaib Attaiki, Gautam Pai, and Maks Ovsjanikov.
\newblock {DPFM}: {D}eep partial functional maps.
\newblock In \emph{2021 International Conference on 3D Vision (3DV)}, pp.\
  175--185. IEEE, 2021.

\bibitem[Atz et~al.(2021)Atz, Grisoni, and Schneider]{At:2021geometric}
Kenneth Atz, Francesca Grisoni, and Gisbert Schneider.
\newblock Geometric deep learning on molecular representations.
\newblock \emph{Nature Machine Intelligence}, 3\penalty0 (12):\penalty0
  1023--1032, 2021.

\bibitem[Aubry et~al.(2011)Aubry, Schlickewei, and Cremers]{aubry2011wave}
Mathieu Aubry, Ulrich Schlickewei, and Daniel Cremers.
\newblock The wave kernel signature: A quantum mechanical approach to shape
  analysis.
\newblock In \emph{2011 IEEE international conference on computer vision
  workshops (ICCV workshops)}, pp.\  1626--1633. IEEE, 2011.

\bibitem[Basu \& Wallner(2016)Basu and Wallner]{basu_dockq_2016}
Sankar Basu and Björn Wallner.
\newblock {DockQ}: {A} {Quality} {Measure} for {Protein}-{Protein} {Docking}
  {Models}.
\newblock \emph{PLOS ONE}, 11\penalty0 (8):\penalty0 e0161879, 2016.

\bibitem[Battiston et~al.(2020)Battiston, Cencetti, Iacopini, Latora, Lucas,
  Patania, Young, and Petri]{battiston2020networks}
Federico Battiston, Giulia Cencetti, Iacopo Iacopini, Vito Latora, Maxime
  Lucas, Alice Patania, Jean-Gabriel Young, and Giovanni Petri.
\newblock Networks beyond pairwise interactions: structure and dynamics.
\newblock \emph{Physics Reports}, 874:\penalty0 1--92, 2020.

\bibitem[Biasotti et~al.(2016)Biasotti, Cerri, Bronstein, and
  Bronstein]{Bi:2016recent}
Silvia Biasotti, Andrea Cerri, Alex Bronstein, and Michael Bronstein.
\newblock Recent trends, applications, and perspectives in 3d shape similarity
  assessment.
\newblock In \emph{Computer graphics forum}, volume~35, pp.\  87--119. Wiley
  Online Library, 2016.

\bibitem[Boguna et~al.(2021)Boguna, Bonamassa, De~Domenico, Havlin, Krioukov,
  and Serrano]{boguna2021network}
Marian Boguna, Ivan Bonamassa, Manlio De~Domenico, Shlomo Havlin, Dmitri
  Krioukov, and M~Serrano.
\newblock Network geometry.
\newblock \emph{Nature Reviews Physics}, 3\penalty0 (2):\penalty0 114--135,
  2021.

\bibitem[Brandstetter et~al.(2021)Brandstetter, Hesselink, van~der Pol,
  Bekkers, and Welling]{brandstetter2021geometric}
Johannes Brandstetter, Rob Hesselink, Elise van~der Pol, Erik Bekkers, and Max
  Welling.
\newblock Geometric and physical quantities improve e (3) equivariant message
  passing.
\newblock \emph{arXiv preprint arXiv:2110.02905}, 2021.

\bibitem[Bronstein \& Kokkinos(2010)Bronstein and Kokkinos]{bronstein2010scale}
Michael~M Bronstein and Iasonas Kokkinos.
\newblock Scale-invariant heat kernel signatures for non-rigid shape
  recognition.
\newblock In \emph{2010 IEEE computer society conference on computer vision and
  pattern recognition}, pp.\  1704--1711. IEEE, 2010.

\bibitem[Bronstein et~al.(2017)Bronstein, Bruna, LeCun, Szlam, and
  Vandergheynst]{Br:2017geometric}
Michael~M Bronstein, Joan Bruna, Yann LeCun, Arthur Szlam, and Pierre
  Vandergheynst.
\newblock Geometric deep learning: going beyond {E}uclidean data.
\newblock \emph{IEEE Signal Processing Magazine}, 34\penalty0 (4):\penalty0
  18--42, 2017.

\bibitem[Bronstein et~al.(2021)Bronstein, Bruna, Cohen, and
  Veli{\v{c}}kovi{\'c}]{Br:2021geometric}
Michael~M Bronstein, Joan Bruna, Taco Cohen, and Petar Veli{\v{c}}kovi{\'c}.
\newblock Geometric deep learning: Grids, groups, graphs, geodesics, and
  gauges.
\newblock \emph{arXiv preprint arXiv:2104.13478}, 2021.

\bibitem[Cohen et~al.(2018)Cohen, Geiger, K{\"o}hler, and
  Welling]{Co:2018spherical}
Taco~S Cohen, Mario Geiger, Jonas K{\"o}hler, and Max Welling.
\newblock Spherical {CNN}s.
\newblock \emph{arXiv preprint arXiv:1801.10130}, 2018.

\bibitem[Coifman \& Lafon(2006)Coifman and Lafon]{coifman2006diffusion}
Ronald~R Coifman and St{\'e}phane Lafon.
\newblock Diffusion maps.
\newblock \emph{Applied and computational harmonic analysis}, 21\penalty0
  (1):\penalty0 5--30, 2006.

\bibitem[Coifman et~al.(2005)Coifman, Lafon, Lee, Maggioni, Nadler, Warner, and
  Zucker]{coifman2005geometric}
Ronald~R Coifman, Stephane Lafon, Ann~B Lee, Mauro Maggioni, Boaz Nadler,
  Frederick Warner, and Steven~W Zucker.
\newblock Geometric diffusions as a tool for harmonic analysis and structure
  definition of data: Diffusion maps.
\newblock \emph{Proceedings of the national academy of sciences}, 102\penalty0
  (21):\penalty0 7426--7431, 2005.

\bibitem[de~Vries et~al.(2015)de~Vries, Schindler, Chauvot~de Beauchêne, and
  Zacharias]{de_vries_attract_2015}
Sjoerd~J. de~Vries, Christina E.~M. Schindler, Isaure Chauvot~de Beauchêne,
  and Martin Zacharias.
\newblock A {Web} {Interface} for {Easy} {Flexible} {Protein}-{Protein}
  {Docking} with {ATTRACT}.
\newblock \emph{Biophysical Journal}, 108\penalty0 (3):\penalty0 462--465,
  February 2015.

\bibitem[Dolinsky et~al.(2004)Dolinsky, Nielsen, McCammon, and
  Baker]{dolinsky2004pdb2pqr}
Todd~J Dolinsky, Jens~E Nielsen, J~Andrew McCammon, and Nathan~A Baker.
\newblock Pdb2pqr: an automated pipeline for the setup of poisson--boltzmann
  electrostatics calculations.
\newblock \emph{Nucleic acids research}, 32\penalty0 (suppl\_2):\penalty0
  W665--W667, 2004.

\bibitem[Donati et~al.(2020)Donati, Sharma, and Ovsjanikov]{Do:2020deep}
Nicolas Donati, Abhishek Sharma, and Maks Ovsjanikov.
\newblock Deep geometric functional maps: {R}obust feature learning for shape
  correspondence.
\newblock In \emph{Proceedings of the IEEE/CVF Conference on Computer Vision
  and Pattern Recognition}, pp.\  8592--8601, 2020.

\bibitem[Duhovny et~al.(2002)Duhovny, Nussinov, and
  Wolfson]{duhovny2002efficient}
Dina Duhovny, Ruth Nussinov, and Haim~J Wolfson.
\newblock Efficient unbound docking of rigid molecules.
\newblock In \emph{International workshop on algorithms in bioinformatics},
  pp.\  185--200. Springer, 2002.

\bibitem[Ewing \& Hermisson(2010)Ewing and Hermisson]{ewing2010msms}
Gregory Ewing and Joachim Hermisson.
\newblock Msms: a coalescent simulation program including recombination,
  demographic structure and selection at a single locus.
\newblock \emph{Bioinformatics}, 26\penalty0 (16):\penalty0 2064--2065, 2010.

\bibitem[Fuchs et~al.(2020)Fuchs, Worrall, Fischer, and Welling]{Fu:2020se}
Fabian Fuchs, Daniel Worrall, Volker Fischer, and Max Welling.
\newblock {SE(3)}-{T}ransformers: 3{D} roto-translation equivariant attention
  networks.
\newblock \emph{Advances in Neural Information Processing Systems},
  33:\penalty0 1970--1981, 2020.

\bibitem[Gainza et~al.(2020)Gainza, Sverrisson, Monti, Rodola, Boscaini,
  Bronstein, and Correia]{Ga:2020deciphering}
Pablo Gainza, Freyr Sverrisson, Frederico Monti, Emanuele Rodola, D~Boscaini,
  MM~Bronstein, and BE~Correia.
\newblock Deciphering interaction fingerprints from protein molecular surfaces
  using geometric deep learning.
\newblock \emph{Nature Methods}, 17\penalty0 (2):\penalty0 184--192, 2020.

\bibitem[Ganea et~al.(2021)Ganea, Huang, Bunne, Bian, Barzilay, Jaakkola, and
  Krause]{Ga:2021independent}
Octavian-Eugen Ganea, Xinyuan Huang, Charlotte Bunne, Yatao Bian, Regina
  Barzilay, Tommi Jaakkola, and Andreas Krause.
\newblock Independent {SE(3)}-equivariant models for end-to-end rigid protein
  docking.
\newblock \emph{arXiv preprint arXiv:2111.07786}, 2021.

\bibitem[Gilmer et~al.(2017)Gilmer, Schoenholz, Riley, Vinyals, and
  Dahl]{Gi:2017neural}
Justin Gilmer, Samuel~S Schoenholz, Patrick~F Riley, Oriol Vinyals, and
  George~E Dahl.
\newblock Neural message passing for quantum chemistry.
\newblock In \emph{International conference on machine learning}, pp.\
  1263--1272. PMLR, 2017.

\bibitem[Gligorijevi{\'c} et~al.(2021)Gligorijevi{\'c}, Renfrew, Kosciolek,
  Leman, Berenberg, Vatanen, Chandler, Taylor, Fisk, Vlamakis,
  et~al.]{Gl:2021structure}
Vladimir Gligorijevi{\'c}, P~Douglas Renfrew, Tomasz Kosciolek, Julia~Koehler
  Leman, Daniel Berenberg, Tommi Vatanen, Chris Chandler, Bryn~C Taylor, Ian~M
  Fisk, Hera Vlamakis, et~al.
\newblock Structure-based protein function prediction using graph convolutional
  networks.
\newblock \emph{Nature communications}, 12\penalty0 (1):\penalty0 1--14, 2021.

\bibitem[Guest et~al.(2021)Guest, Vreven, Zhou, Moal, Jeliazkov, Gray, Weng,
  and Pierce]{guest_expanded_db55_2021}
Johnathan~D. Guest, Thom Vreven, Jing Zhou, Iain Moal, Jeliazko~R. Jeliazkov,
  Jeffrey~J. Gray, Zhiping Weng, and Brian~G. Pierce.
\newblock An expanded benchmark for antibody-antigen docking and affinity
  prediction reveals insights into antibody recognition determinants.
\newblock \emph{Structure}, 29\penalty0 (6):\penalty0 606--621.e5, June 2021.

\bibitem[Isert et~al.(2022)Isert, Atz, and Schneider]{isert2022structure}
Clemens Isert, Kenneth Atz, and Gisbert Schneider.
\newblock Structure-based drug design with geometric deep learning.
\newblock \emph{arXiv preprint arXiv:2210.11250}, 2022.

\bibitem[Jacobson \& Panozzo(2017)Jacobson and Panozzo]{jacobson2017libigl}
Alec Jacobson and Daniele Panozzo.
\newblock Libigl: Prototyping geometry processing research in c++.
\newblock In \emph{SIGGRAPH Asia 2017 courses}, pp.\  1--172. 2017.

\bibitem[Jumper et~al.(2021)Jumper, Evans, Pritzel, Green, Figurnov,
  Ronneberger, Tunyasuvunakool, Bates, {\v{Z}}{\'\i}dek, Potapenko,
  et~al.]{jumper2021highly}
John Jumper, Richard Evans, Alexander Pritzel, Tim Green, Michael Figurnov,
  Olaf Ronneberger, Kathryn Tunyasuvunakool, Russ Bates, Augustin
  {\v{Z}}{\'\i}dek, Anna Potapenko, et~al.
\newblock Highly accurate protein structure prediction with alphafold.
\newblock \emph{Nature}, 596\penalty0 (7873):\penalty0 583--589, 2021.

\bibitem[Kabsch(1976)]{Ka:1976solution}
Wolfgang Kabsch.
\newblock A solution for the best rotation to relate two sets of vectors.
\newblock \emph{Acta Crystallographica Section A: Crystal Physics, Diffraction,
  Theoretical and General Crystallography}, 32\penalty0 (5):\penalty0 922--923,
  1976.

\bibitem[Kac(1966)]{kac1966can}
Mark Kac.
\newblock Can one hear the shape of a drum?
\newblock \emph{The american mathematical monthly}, 73\penalty0 (4P2):\penalty0
  1--23, 1966.

\bibitem[Kipf \& Welling(2016)Kipf and Welling]{kipf2016semi}
Thomas~N Kipf and Max Welling.
\newblock Semi-supervised classification with graph convolutional networks.
\newblock \emph{arXiv preprint arXiv:1609.02907}, 2016.

\bibitem[Klicpera et~al.(2020)Klicpera, Gro{\ss}, and
  G{\"u}nnemann]{Kl:2020directional}
Johannes Klicpera, Janek Gro{\ss}, and Stephan G{\"u}nnemann.
\newblock Directional message passing for molecular graphs.
\newblock \emph{arXiv preprint arXiv:2003.03123}, 2020.

\bibitem[Lee(2013)]{lee2013smooth}
John~M Lee.
\newblock Smooth manifolds.
\newblock In \emph{Introduction to smooth manifolds}, pp.\  1--31. Springer,
  2013.

\bibitem[Lensink \& Wodak(2013)Lensink and Wodak]{lensink_docking_capri_2013}
Marc~F. Lensink and Shoshana~J. Wodak.
\newblock Docking, scoring, and affinity prediction in {CAPRI}.
\newblock \emph{Proteins: Structure, Function, and Bioinformatics}, 81\penalty0
  (12):\penalty0 2082--2095, 2013.

\bibitem[Li et~al.(2021)Li, Li, Chen, Fu, Cohen-Or, and Heng]{Li:2021rotation}
Xianzhi Li, Ruihui Li, Guangyong Chen, Chi-Wing Fu, Daniel Cohen-Or, and
  Pheng-Ann Heng.
\newblock A rotation-invariant framework for deep point cloud analysis.
\newblock \emph{IEEE Transactions on Visualization and Computer Graphics},
  2021.

\bibitem[Li et~al.(2013)Li, Zhang, and Cao]{Li:2013role}
Ye~Li, Xianren Zhang, and Dapeng Cao.
\newblock The role of shape complementarity in the protein-protein
  interactions.
\newblock \emph{Scientific reports}, 3\penalty0 (1):\penalty0 1--7, 2013.

\bibitem[Litany et~al.(2017{\natexlab{a}})Litany, Remez, Rodola, Bronstein, and
  Bronstein]{Li:2017deep}
Or~Litany, Tal Remez, Emanuele Rodola, Alex Bronstein, and Michael Bronstein.
\newblock Deep functional maps: {S}tructured prediction for dense shape
  correspondence.
\newblock In \emph{Proceedings of the IEEE international conference on computer
  vision}, pp.\  5659--5667, 2017{\natexlab{a}}.

\bibitem[Litany et~al.(2017{\natexlab{b}})Litany, Rodol{\`a}, Bronstein, and
  Bronstein]{litany2017fully}
Or~Litany, Emanuele Rodol{\`a}, Alexander~M Bronstein, and Michael~M Bronstein.
\newblock Fully spectral partial shape matching.
\newblock In \emph{Computer Graphics Forum}, volume~36, pp.\  247--258. Wiley
  Online Library, 2017{\natexlab{b}}.

\bibitem[Litman \& Bronstein(2013)Litman and Bronstein]{Li:2013learning}
Roee Litman and Alexander~M Bronstein.
\newblock Learning spectral descriptors for deformable shape correspondence.
\newblock \emph{IEEE transactions on pattern analysis and machine
  intelligence}, 36\penalty0 (1):\penalty0 171--180, 2013.

\bibitem[Liu et~al.(2021)Liu, Wang, Zhu, Gaines, Zhu, Bi, and
  Song]{Li:2021octsurf}
Qinqing Liu, Peng-Shuai Wang, Chunjiang Zhu, Blake~Blumenfeld Gaines, Tan Zhu,
  Jinbo Bi, and Minghu Song.
\newblock Oct{S}urf: Efficient hierarchical voxel-based molecular surface
  representation for protein-ligand affinity prediction.
\newblock \emph{Journal of Molecular Graphics and Modelling}, 105:\penalty0
  107865, 2021.

\bibitem[Monti et~al.(2017)Monti, Boscaini, Masci, Rodola, Svoboda, and
  Bronstein]{Mo:2017geometric}
Federico Monti, Davide Boscaini, Jonathan Masci, Emanuele Rodola, Jan Svoboda,
  and Michael~M Bronstein.
\newblock Geometric deep learning on graphs and manifolds using mixture model
  {CNN}s.
\newblock In \emph{Proceedings of the IEEE conference on computer vision and
  pattern recognition}, pp.\  5115--5124, 2017.

\bibitem[Mylonas et~al.(2021)Mylonas, Axenopoulos, and Daras]{My:2021deepsurf}
Stelios~K Mylonas, Apostolos Axenopoulos, and Petros Daras.
\newblock Deep{S}urf: a surface-based deep learning approach for the prediction
  of ligand binding sites on proteins.
\newblock \emph{Bioinformatics}, 37\penalty0 (12):\penalty0 1681--1690, 2021.

\bibitem[Méndez et~al.(2003)Méndez, Leplae, De~Maria, and
  Wodak]{mendez_assessment_2003}
Raúl Méndez, Raphaël Leplae, Leonardo De~Maria, and Shoshana~J. Wodak.
\newblock Assessment of blind predictions of protein–protein interactions:
  {Current} status of docking methods.
\newblock \emph{Proteins: Structure, Function, and Bioinformatics}, 52\penalty0
  (1):\penalty0 51--67, 2003.

\bibitem[Méndez et~al.(2005)Méndez, Leplae, Lensink, and
  Wodak]{mendez_assessment_2005}
Raúl Méndez, Raphaël Leplae, Marc~F. Lensink, and Shoshana~J. Wodak.
\newblock Assessment of {CAPRI} predictions in rounds 3–5 shows progress in
  docking procedures.
\newblock \emph{Proteins: Structure, Function, and Bioinformatics}, 60\penalty0
  (2):\penalty0 150--169, 2005.

\bibitem[Ovsjanikov et~al.(2012)Ovsjanikov, Ben-Chen, Solomon, Butscher, and
  Guibas]{Ov:2012functional}
Maks Ovsjanikov, Mirela Ben-Chen, Justin Solomon, Adrian Butscher, and Leonidas
  Guibas.
\newblock Functional maps: a flexible representation of maps between shapes.
\newblock \emph{ACM Transactions on Graphics (ToG)}, 31\penalty0 (4):\penalty0
  1--11, 2012.

\bibitem[Petersen(2006)]{petersen2006riemannian}
Peter Petersen.
\newblock \emph{Riemannian geometry}, volume 171.
\newblock Springer, 2006.

\bibitem[Ramakrishnan et~al.(2014)Ramakrishnan, Dral, Rupp, and
  Von~Lilienfeld]{Ra:2014quantum}
Raghunathan Ramakrishnan, Pavlo~O Dral, Matthias Rupp, and O~Anatole
  Von~Lilienfeld.
\newblock Quantum chemistry structures and properties of 134 kilo molecules.
\newblock \emph{Scientific data}, 1\penalty0 (1):\penalty0 1--7, 2014.

\bibitem[Ren et~al.(2018)Ren, Poulenard, Wonka, and
  Ovsjanikov]{Re:2018continuous}
Jing Ren, Adrien Poulenard, Peter Wonka, and Maks Ovsjanikov.
\newblock Continuous and orientation-preserving correspondences via functional
  maps.
\newblock \emph{ACM Transactions on Graphics (ToG)}, 37\penalty0 (6):\penalty0
  1--16, 2018.

\bibitem[Reuter et~al.(2006)Reuter, Wolter, and Peinecke]{reuter2006laplace}
Martin Reuter, Franz-Erich Wolter, and Niklas Peinecke.
\newblock Laplace--beltrami spectra as ‘shape-dna’of surfaces and solids.
\newblock \emph{Computer-Aided Design}, 38\penalty0 (4):\penalty0 342--366,
  2006.

\bibitem[Reuter et~al.(2009)Reuter, Biasotti, Giorgi, Patan{\`e}, and
  Spagnuolo]{reuter2009discrete}
Martin Reuter, Silvia Biasotti, Daniela Giorgi, Giuseppe Patan{\`e}, and
  Michela Spagnuolo.
\newblock Discrete laplace--beltrami operators for shape analysis and
  segmentation.
\newblock \emph{Computers \& Graphics}, 33\penalty0 (3):\penalty0 381--390,
  2009.

\bibitem[Richards(1977)]{richards1977areas}
Frederic~M Richards.
\newblock Areas, volumes, packing, and protein structure.
\newblock \emph{Annual review of biophysics and bioengineering}, 6\penalty0
  (1):\penalty0 151--176, 1977.

\bibitem[Rosenberg(1997)]{Ro:1997laplacian}
Steven Rosenberg.
\newblock \emph{The Laplacian on a Riemannian manifold: an introduction to
  analysis on manifolds}.
\newblock Number~31. Cambridge University Press, 1997.

\bibitem[Satorras et~al.(2021)Satorras, Hoogeboom, and Welling]{Sa:2021n}
V{\i}ctor~Garcia Satorras, Emiel Hoogeboom, and Max Welling.
\newblock E(n) equivariant graph neural networks.
\newblock In \emph{International conference on machine learning}, pp.\
  9323--9332. PMLR, 2021.

\bibitem[Sch{\"u}tt et~al.(2018)Sch{\"u}tt, Sauceda, Kindermans, Tkatchenko,
  and M{\"u}ller]{schutt2018schnet}
Kristof~T Sch{\"u}tt, Huziel~E Sauceda, P-J Kindermans, Alexandre Tkatchenko,
  and K-R M{\"u}ller.
\newblock Schnet--a deep learning architecture for molecules and materials.
\newblock \emph{The Journal of Chemical Physics}, 148\penalty0 (24):\penalty0
  241722, 2018.

\bibitem[Sharp(1994)]{sharp1994electrostatic}
Kim~A Sharp.
\newblock Electrostatic interactions in macromolecules.
\newblock \emph{Current Opinion in Structural Biology}, 4\penalty0
  (2):\penalty0 234--239, 1994.

\bibitem[Shen et~al.(2021)Shen, Dai, Li, Zou, and Xiong]{shen2021multi}
Yangmei Shen, Wenrui Dai, Chenglin Li, Junni Zou, and Hongkai Xiong.
\newblock Multi-scale graph convolutional network with spectral graph wavelet
  frame.
\newblock \emph{IEEE Transactions on Signal and Information Processing over
  Networks}, 7:\penalty0 595--610, 2021.

\bibitem[Shulman-Peleg et~al.(2004)Shulman-Peleg, Nussinov, and
  Wolfson]{shulman2004recognition}
Alexandra Shulman-Peleg, Ruth Nussinov, and Haim~J Wolfson.
\newblock Recognition of functional sites in protein structures.
\newblock \emph{Journal of molecular biology}, 339\penalty0 (3):\penalty0
  607--633, 2004.

\bibitem[Somnath et~al.(2021)Somnath, Bunne, and Krause]{somnath2021multi}
Vignesh~Ram Somnath, Charlotte Bunne, and Andreas Krause.
\newblock Multi-scale representation learning on proteins.
\newblock \emph{Advances in Neural Information Processing Systems},
  34:\penalty0 25244--25255, 2021.

\bibitem[Stokes et~al.(2020)Stokes, Yang, Swanson, Jin, Cubillos-Ruiz, Donghia,
  MacNair, French, Carfrae, Bloom-Ackermann, et~al.]{St:2020deep}
Jonathan~M Stokes, Kevin Yang, Kyle Swanson, Wengong Jin, Andres Cubillos-Ruiz,
  Nina~M Donghia, Craig~R MacNair, Shawn French, Lindsey~A Carfrae, Zohar
  Bloom-Ackermann, et~al.
\newblock A deep learning approach to antibiotic discovery.
\newblock \emph{Cell}, 180\penalty0 (4):\penalty0 688--702, 2020.

\bibitem[Sun et~al.(2009)Sun, Ovsjanikov, and Guibas]{sun2009concise}
Jian Sun, Maks Ovsjanikov, and Leonidas Guibas.
\newblock A concise and provably informative multi-scale signature based on
  heat diffusion.
\newblock In \emph{Computer graphics forum}, volume~28, pp.\  1383--1392. Wiley
  Online Library, 2009.

\bibitem[Sverrisson et~al.(2021)Sverrisson, Feydy, Correia, and
  Bronstein]{sverrisson2021fast}
Freyr Sverrisson, Jean Feydy, Bruno~E Correia, and Michael~M Bronstein.
\newblock Fast end-to-end learning on protein surfaces.
\newblock In \emph{Proceedings of the IEEE/CVF Conference on Computer Vision
  and Pattern Recognition}, pp.\  15272--15281, 2021.

\bibitem[Sverrisson et~al.(2022)Sverrisson, Feydy, Southern, Bronstein, and
  Correia]{Sv:2022physics}
Freyr Sverrisson, Jean Feydy, Joshua Southern, Michael~M Bronstein, and Bruno
  Correia.
\newblock Physics-informed deep neural network for rigid-body protein docking.
\newblock In \emph{ICLR2022 Machine Learning for Drug Discovery}, 2022.

\bibitem[Thomas et~al.(2018)Thomas, Smidt, Kearnes, Yang, Li, Kohlhoff, and
  Riley]{Th:2018tensor}
Nathaniel Thomas, Tess Smidt, Steven Kearnes, Lusann Yang, Li~Li, Kai Kohlhoff,
  and Patrick Riley.
\newblock Tensor field networks: Rotation-and translation-equivariant neural
  networks for 3d point clouds.
\newblock \emph{arXiv preprint arXiv:1802.08219}, 2018.

\bibitem[Townshend et~al.(2019)Townshend, Bedi, Suriana, and
  Dror]{townshend_dips_2019}
Raphael J.~L. Townshend, Rishi Bedi, Patricia~A. Suriana, and Ron~O. Dror.
\newblock End-to-{End} {Learning} on {3D} {Protein} {Structure} for {Interface}
  {Prediction}.
\newblock Technical report, arXiv, December 2019.
\newblock arXiv:1807.01297 [cs, q-bio, stat] type: article.

\bibitem[Townshend et~al.(2021)Townshend, Eismann, Watkins, Rangan, Karelina,
  Das, and Dror]{townshend2021geometric}
Raphael~JL Townshend, Stephan Eismann, Andrew~M Watkins, Ramya Rangan, Maria
  Karelina, Rhiju Das, and Ron~O Dror.
\newblock Geometric deep learning of rna structure.
\newblock \emph{Science}, 373\penalty0 (6558):\penalty0 1047--1051, 2021.

\bibitem[Unke et~al.(2021)Unke, Bogojeski, Gastegger, Geiger, Smidt, and
  M{\"u}ller]{Un:2021se}
Oliver Unke, Mihail Bogojeski, Michael Gastegger, Mario Geiger, Tess Smidt, and
  Klaus-Robert M{\"u}ller.
\newblock {SE}(3)-equivariant prediction of molecular wavefunctions and
  electronic densities.
\newblock \emph{Advances in Neural Information Processing Systems},
  34:\penalty0 14434--14447, 2021.

\bibitem[Veličković et~al.(2018)Veličković, Cucurull, Casanova, Romero,
  Liò, and Bengio]{velickovic_graph_2018}
Petar Veličković, Guillem Cucurull, Arantxa Casanova, Adriana Romero, Pietro
  Liò, and Yoshua Bengio.
\newblock Graph {Attention} {Networks}, February 2018.
\newblock URL \url{http://arxiv.org/abs/1710.10903}.
\newblock arXiv:1710.10903 [cs, stat].

\bibitem[Word et~al.(1999)Word, Lovell, Richardson, and
  Richardson]{word1999asparagine}
J~Michael Word, Simon~C Lovell, Jane~S Richardson, and David~C Richardson.
\newblock Asparagine and glutamine: using hydrogen atom contacts in the choice
  of side-chain amide orientation.
\newblock \emph{Journal of molecular biology}, 285\penalty0 (4):\penalty0
  1735--1747, 1999.

\bibitem[Xie et~al.(2020)Xie, Gu, Guo, Qi, Guibas, and
  Litany]{xie2020pointcontrast}
Saining Xie, Jiatao Gu, Demi Guo, Charles~R Qi, Leonidas Guibas, and Or~Litany.
\newblock Pointcontrast: Unsupervised pre-training for 3d point cloud
  understanding.
\newblock In \emph{European conference on computer vision}, pp.\  574--591.
  Springer, 2020.

\bibitem[Yan et~al.(2020)Yan, Tao, He, and Huang]{yan_hdock_2020}
Yumeng Yan, Huanyu Tao, Jiahua He, and Sheng-You Huang.
\newblock The {HDOCK} server for integrated protein–protein docking.
\newblock \emph{Nature Protocols}, 15\penalty0 (5):\penalty0 1829--1852, May
  2020.
\newblock \doi{10.1038/s41596-020-0312-x}.
\newblock Number: 5 Publisher: Nature Publishing Group.

\bibitem[Zhou(2019)]{zhou2019pymesh}
Q~Zhou.
\newblock Pymesh—geometry processing library for python.
\newblock \emph{Software available for download at
  https://github.com/PyMesh/PyMesh}, 2019.

\end{thebibliography}
